\documentclass[runningheads]{llncs}
\usepackage{graphicx}
\usepackage{url}

\usepackage{amsfonts}
\usepackage{amsmath}
\usepackage[noend,algo2e,boxruled,linesnumbered]{algorithm2e}
\usepackage{bbm}
\usepackage{type1cm}
\usepackage{xcolor}
\usepackage{tabularx}
\usepackage{booktabs}
\usepackage{multirow}
\usepackage{subcaption}

\newcommand{\ncd}{\textsc{ncd}}
\newcommand{\ncda}{\textsc{ncda}}
\newcommand{\gencda}{\textsc{ge\ncda{}}}
\newcommand{\lime}{\textsc{lime}}
\newcommand{\calime}{\textsc{ca\lime{}}}

\newcommand{\banknote}{\texttt{banknote}}
\newcommand{\wdbc}{\texttt{wdbc}}
\newcommand{\statlog}{\texttt{statlog}}

\newcommand{\magic}{\texttt{magic}}
\newcommand{\wine}{\texttt{wine-red}}
\begin{document}
%
\title{Causality-Aware Local Interpretable Model-Agnostic Explanations}

\titlerunning{CALIME}

\author{Martina Cinquini \and Riccardo Guidotti}

\authorrunning{Martina Cinquini \& Riccardo Guidotti}

\tocauthor{Martina Cinquini, Riccardo Guidotti}

\institute{University of Pisa, Largo B. Pontecorvo, 3, 56127, Pisa, Italy. \email{martina.cinquini@phd.unipi.it, riccardo.guidotti@unipi.it}}
\maketitle              
\begin{abstract}
A main drawback of eXplainable Artificial Intelligence (XAI) approaches is the feature independence assumption, hindering the study of potential variable dependencies.
This leads to approximating black box behaviors by analyzing the effects on randomly generated feature values that may rarely occur in the original samples. 
This paper addresses this issue by integrating causal knowledge in an XAI method to enhance transparency and enable users to assess the quality of the generated explanations.
Specifically, we propose a novel extension to a widely used local and model-agnostic explainer, which encodes explicit causal relationships within the data surrounding the instance being explained.
Extensive experiments show that our approach overcomes the original method in terms of faithfully replicating the black-box model's mechanism and the consistency and reliability of the generated explanations.


\end{abstract}

\section{Introduction}
\label{sec:intro}
In past decades, the growing availability of data and computational power have allowed the development of sophisticated Machine Learning (ML) models that have provided a significant contribution to the progress of Artificial Intelligence (AI) in many real-world applications~\cite{preece2018asking}.
Despite their success, the surge in performance has often been achieved through increased model complexity, turning such approaches into black-box systems. 
The lack of a clear interpretation of the internal structure of ML models 
embodies a crucial weakness since it causes uncertainty regarding the way they operate and, ultimately, how they infer outcomes~\cite{linardatos2021explainable}. 
Indeed, opaque approaches may inadvertently obfuscate responsibility for any biases or produce wrong decisions learned from spurious correlations in training data~\cite{guidotti2019survey}.
To mitigate these risks, the scientific community's interest in eXplainable Artificial Intelligence (XAI) has been increasingly emerging in the past decade. 
According to the classification of XAI methodologies~\cite{guidotti2019survey}, we focus on post-hoc explainability.
Given an AI decision system based on a black-box classifier and an instance to explain, post-hoc local explanation methods approximate the black-box behavior by learning an interpretable model in a synthetic neighborhood of the instance under analysis generated randomly.
However, particular combinations of feature values might be unrealistic, leading to implausible synthetic instances~\cite{artelt2021evaluating,laugel2019dangers}.
This weakness emerges since these methods do not consider the local distribution features, the density of the class labels in the neighborhood~\cite{moradi2021post}, and, most importantly, the causal relationships among input features~\cite{longo2020explainable}.
Such inability to disentangle correlation from causation can deliver sub-optimal or even erroneous explanations to decision-makers~\cite{richens2020improving}. 
Moreover, causation is ubiquitous in Humans’ conception of their environment~\cite{chou2022counterfactuals}. 
Human beings are extremely good at constructing mental decision models from a few data samples because people excel at generalizing data and thinking in a cause/effect manner. 
Consequently, integrating \textit{causal knowledge} in XAI should be emphasized to attain a higher degree of interpretability and prevent procedures from failing in unexpected circumstances~\cite{beretta2023importance}. 
To this aim, Causal Discovery (CD) methods can map all the causal pathways to a variable and infer how different variables are related. 
Even partial knowledge of the causal structure of observational data could improve understanding of which input features black-box models have used to make their predictions, allowing for a higher degree of interpretability and more robust explanations.

In this paper, we propose \calime{} for Causality-Aware Local Interpretable Model-Agnostic Explanations. 
The method extends \lime{}~\cite{ribeiro2016should} by accounting for underlying causal relationships present in the data used by the black-box. 
To attain this purpose, we replace the synthetic data generation performed by \lime{} through random sampling with \gencda{}~\cite{cinquini2021gencda}, a synthetic dataset generator for tabular data that explicitly allows for encoding causal dependencies.
We would like to emphasize that causal relationships are not typically known a priori, but we adopt a CD approach to discover nonlinear causalities among the variables and use them at generation time.
Thus, the novelty of \calime{} lies in the \textit{explicit encoding of causal relationships} in the local synthetic neighborhood of the instance for which the explanation is required.
We highlight that our proposal of the causality-aware explanation method can be easily adapted to extend and improve other model-agnostic explainers such as \textsc{lore}~\cite{guidotti2019factual}, or \textsc{shap}~\cite{lundberg2017unified}.
Our experiments show that \calime{} significantly improves over \lime{} both for the stability of explanations and the fidelity in mimicking the black-box. 

The rest of this paper is organized as follows. 
Section~\ref{sec:related} describes the state-of-the-art related to XAI and causality. 
Section~\ref{sec:background} presents a formalization of the problem addressed and recalls basic concepts for understanding our proposal, which is detailed in Section~\ref{sec:method}.
The experimental results are presented in Section~\ref{sec:experiments}, while the conclusions, as well as future works, are discussed in Section~\ref{sec:conclusion}.

\section{Related Works}\label{sec:related}
\lime{}~\cite{ribeiro2016should} is a model-agnostic method that returns local explanations as feature importance vectors. Further details are in Section~\ref{sec:background} as it is the starting point for our proposal.
While \lime{} stands out for its simplicity and effectiveness, it has
several weak points.
A significant limitation is its reliance on converting all data into a binary format, which is assumed to be more interpretable for humans~\cite{guidotti2019investigating}. Additionally, its use of random perturbations to generate explanations can lead to inconsistencies, possibly producing varying explanations for the same input and prediction across different runs~\cite{zafar2021deterministic}. Furthermore, the reliability of additive explanations comes into question, especially when noisy interactions are introduced in the reference dataset used for generating explanations~\cite{gosiewska2019not}.

Over recent years, numerous researchers have analyzed such limitations and proposed several works to extend or improve them. 
For instance, \textsc{klime}~\cite{hall2017machine} runs the K-Means clustering algorithm to partition the training data and then fit local models within each cluster instead of perturbation-based data generation around an instance being explained. 
In~\cite{hu2018locally} is proposed \textsc{lime-sup} that approximates the original \lime{} better than \textsc{klime} by using supervised partitioning. 
Furthermore, \textsc{kl-lime}~\cite{peltola2018local} adopts the Kullback-Leibler divergence to explain Bayesian predictive models. 
Within this constraint, both the original task and the explanation model can be arbitrarily changed without losing the theoretical information interpretation of the projection for finding the explanation model.
\textsc{alime}~\cite{shankaranarayana2019alime} presented modifications using an autoencoder as a better weighting function for the local surrogate models.
In \textsc{qlime}~\cite{bramhall2020qlime}, the authors consider nonlinear relationships using a quadratic approximation.
Another approach proposed in~\cite{saito2020improving} utilizes a Conditional Tabular Generative Adversarial Network (CTGAN) to generate more realistic synthetic data for querying the model to be explained. 
Theoretically, GAN-like methods can learn possible dependencies.
However, as empirically demonstrated in~\cite{cinquini2021gencda}, these relationships are not directly represented, and there is no guarantee that they are followed in the data generation process.
In~\cite{zafar2021deterministic} is proposed \textsc{dlime}, a Deterministic Local Interpretable Model-Agnostic Explanations where random perturbations are replaced with hierarchical clustering to group the data.
After that, a kNN is used to select the cluster where the instance to be explained belongs. 
Finally, in~\cite{zhao2021baylime} is presented \textsc{bay-lime}, a Bayesian local surrogate model that exploits prior knowledge and Bayesian reasoning to improve both the consistency in repeated explanations of a single prediction and the robustness of kernel settings.
 
Despite considerable progress in \lime{} variants, a significant limitation remains unaddressed in the state-of-the-art approaches: they do not explicitly account for causal relationships. 
This gap represents a fundamental shortfall. Nonetheless, employing a causal framework for deriving explanations could lead to a more reliable and solid explanatory process~\cite{beretta2023importance}. 
Our research demonstrates that local surrogate models exhibit increased fidelity when trained within synthetic environments, taking causality into account.

XAI methods integrating causal knowledge are a recently challenging research area~\cite{moraffah2020causal}.
Among them, one of the most popular methods for Counterfactual Explanation (CE) shows how the prediction result changes with small perturbations to the input~\cite{wachter2017counterfactual}. 
In~\cite{karimi2021algorithmic}, a focus is presented on the role of causality towards feasible counterfactual explanations requiring complete knowledge of the causal model.
Another CE approach called Ordered Counterfactual Explanation is in~\cite{kanamori2021ordered}. It works under the assumption of linear causal relationships and provides an optimal perturbation vector and the order of the features to be perturbed, respecting causal relationships.

Even though the explainers mentioned above account for causal relationships, to the best of our knowledge, no state-of-the-art XAI techniques incorporate any causal knowledge during the explanation extraction. 
Some indirectly account for causality through a latent representation or adopt a known causal graph, typically as a post-hoc filtering step~\cite{parafita2019explaining}. Others consider causal relations between the input features and the outcome label but are not directly interested in the interactions among input features~\cite{joshi2019towards}.

Hence, our method represents a novel post-hoc local explanation strategy by embedding causal connections into the explanation generation process. 
In particular, it does not necessitate pre-existing causal information but independently identifies causal relationships as part of the explanation extraction phase.

\section{Background}
\label{sec:background}
This paper addresses the \textit{black-box outcome explanation problem}~\cite{guidotti2019survey}.
A classifier $b$ is \emph{black-box} if its internal mechanisms are either hidden from the observer or, even if known, they remain incomprehensible to humans.
Given $b$ and an instance $x$ classified by $b$,~i.e., $b(x) = y$, the black-box outcome explanation problem aims at explaining $e$ belonging to a human-interpretable domain.
In our work, we focus on feature importance modeling the explanation as a vector $e = \{ e_1, e_2, \dots,$ $e_m \}$, in which the value 
$e_i \in e$ is the importance of the $i^{\mathit{th}}$ feature for the decision made by $b(x)$. 
To understand the contribution of each feature, the sign and the magnitude of $e_i$ are considered: 
if $e_i < 0$, the feature contributes negatively to the outcome $y$; otherwise, the feature contributes positively. 
The magnitude represents how significant the feature's contribution is to the prediction.

\smallskip
\textbf{LIME.} 
The main idea of \lime{}~\cite{ribeiro2016should} is that the explanation may be derived locally from records generated randomly in the synthetic neighborhood $Z$ of the instance $x$ to be explained. 
\lime{} randomly draws samples and weights them w.r.t. a certain distance function $\pi$ to capture the proximity with $x$. Then, a perturbed sample of instances $Z$ is used to feed to the black-box $b$ and obtain the classification probabilities $b_p(Z)$ w.r.t. the class $b(x) = y$.
Such $b_p(Z)$ are combined with the weights $W$ to train a linear regressor with Lasso regularization considering the top $k$ most essential features. The coefficients of the linear regressor are returned as explanation $e$.
A crucial weakness of \lime{} is the \textit{randomness} of perturbations around the instance to be explained. 
To address this problem, we employ a data generation process that provides more realistic data respecting causal relationships.

\smallskip
\textbf{GENCDA.}
In~\cite{cinquini2021gencda} is presented \gencda{}, a synthetic data generator for tabular data that explicitly allows for encoding the causal structure among variables using \ncda{}, a boosted version of \ncd{}~\cite{hoyer2008nonlinear}. 
Assuming there is no confounding, no selection bias, and no feedback among variables, \ncd{} recovers the causal graph $\mathcal{G}$ from the observational distribution by exploring a functional causal model in which effects are modeled as nonlinear functions of their causes and where the influence of the noise is restricted to be additive.
%
\gencda{} takes as input the \textit{real} dataset $X$ that has to be extended with \textit{synthetic} data, the DAG $\mathcal{G}$ extracted from $X$ by \ncda{} and a set of distributions. 
It generates a synthetic dataset with variable dependencies that adhere to the causal relationships modeled in $\mathcal{G}$.

\begin{figure}[t]
    \centering
    \includegraphics[width=0.80\textwidth]{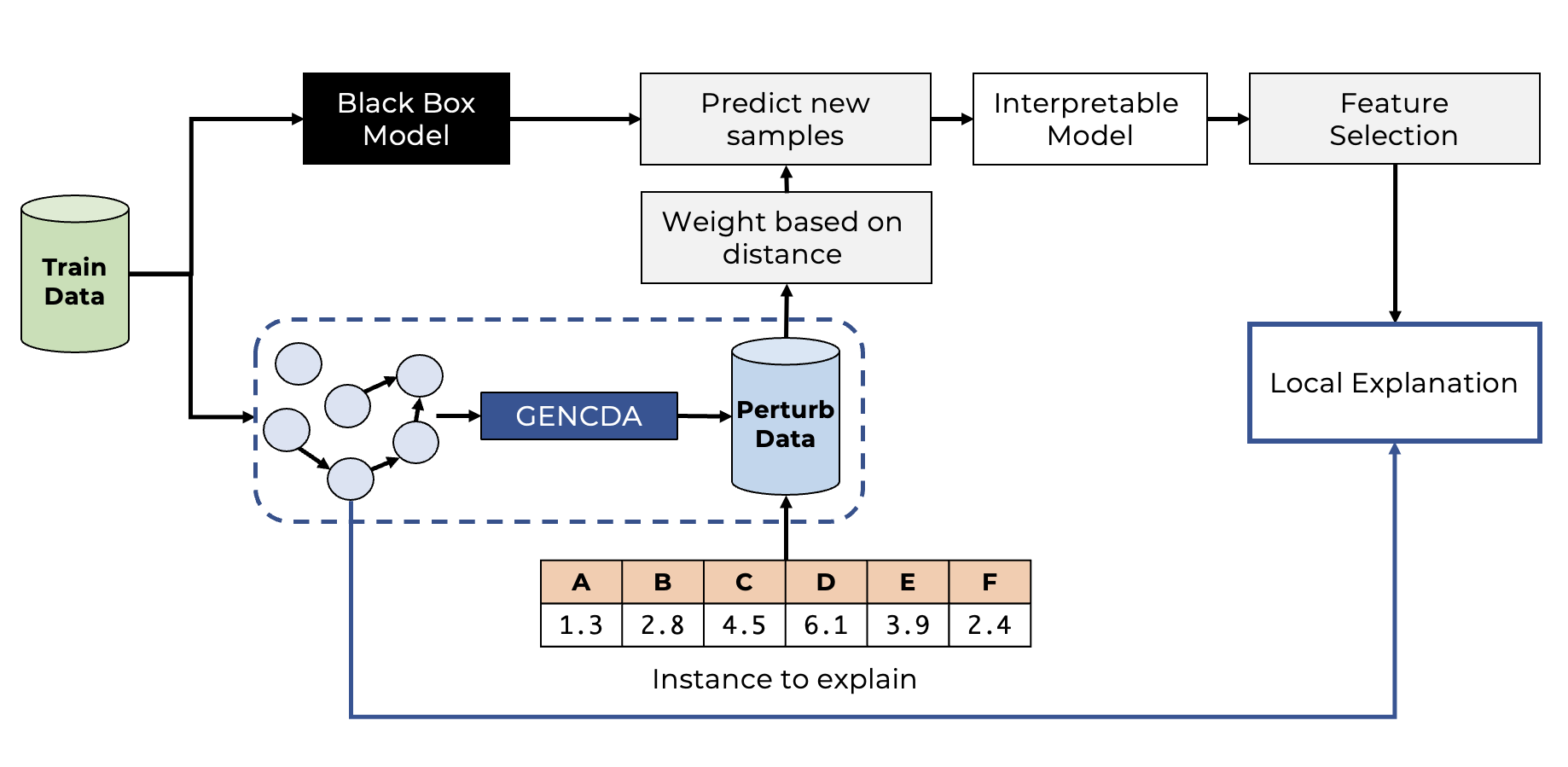}
    \caption{The high-level workflow of our framework, with the blue dashed line showing the difference in generating the synthetic neighborhood compared to \lime{}.}
    \label{fig:calime}
\end{figure}

\section{Causality-Aware LIME}
\label{sec:method}
This section describes our method for Causality-Aware Local Interpretable Model-agnostic Explanations (\calime{}). 
Before introducing it, we provide a practical example to demonstrate the potential limitations of generating explanations without considering causal relations.
Consider a dataset $X$ that describes a bank's customers requesting loans. 
This dataset includes features such as \textit{age}, \textit{education level}, \textit{income}, and \textit{number of weekly working hours}.
Let $b$ be the AI system based on a black-box adopted by the bank and used to grant loans to customers $x \in X$. 
Consider $x = \{(\text{age}, 24), (\text{edu\_level}, \text{high-school-5}), $ $(\text{income}, 800), (\text{work\_hours},$ $20) \}$ a customer who was denied the loan.
To offer additional explanations for the loan denial, the bank decides to apply the \lime{} algorithm to the instance $x$.
A possible explanation for $b(x)$ could be $e = \{(\text{age}, 0.3), (\text{edu\_level}, 0.9), (\text{income}, 0.2), (\text{work\_hours}, 0) \}$.
Thus, according to $e$, it appears that the primary factor contributing to the loan denial is the low \textit{education level}.
By examining the neighborhood $Z$ generated by \lime{}, we may come across several synthetic instances like $z = \{(\text{age}, 24), (\text{edu\_level}, \text{phd}\text{-8}),$ $(\text{income}, 900), (\text{work\_hours}, 20) \}$ where a higher education level is observed.
If we had known the causal relationships among the customers in the training set of the black-box, we would have discovered that there is a 
relationship between \textit{age} and \textit{education level}, i.e., $\text{age} \rightarrow \text{edu\_level}$, and that synthetic instances like $z$ are implausible because they do not respect such relationship.

Therefore, the key idea of \calime{} is to locally explain the predictions of any black-box model while considering the underlying causal relationships within the data. 
The novelty of our proposal is actualized in \textit{the generation of the neighborhood} around the instance to explain. 
Indeed, instead of random perturbation, \calime{} uses \gencda{} as a synthetic dataset generator for tabular data, which can discover nonlinear dependencies among the features and use them during the generation process.
While the improvement of \calime{} over \lime{} may seem straightforward, the ability to encode explicitly causal relationships provides a significant added value to explanations.
In particular, respecting dependencies during the explanation process ensures that local explanations are based on plausible synthetic data and that the final explanations are more trustworthy and less subject to possible noise in the data. Figure \ref{fig:calime} illustrates a high-level workflow depicting our proposal. 

\begin{algorithm2e}[t]
	\footnotesize
    \caption{\calime{}($x$, $b$, $X$, $k$, $N$)}
	\label{alg:calime}
	\SetKwInOut{Input}{Input}
	\SetKwInOut{Output}{Output}
	\Input{$x$ - instance to explain, 
	    $b$ - classifier,
	    $X$ - reference dataset,
	    $k$ - nbr of features,
	    $N$ - nbr of samples
	}
	\Output{$e$ - features importance}
	\BlankLine
	$Z \leftarrow \emptyset, W \leftarrow \emptyset$;\tcp*[f]{\footnotesize \texttt{init. empty synth data and weights}}\\  
	{\color{blue} $G \leftarrow \textsc{ncda}(X)$; \tcp*[f]{\footnotesize \texttt{extract DAG modeling causal relationships}}}\\
	{\color{blue} $I \leftarrow \mathit{fit\_distributions}(G, X)$; \tcp*[f]{\footnotesize \texttt{learn root distributions}}}\\
	{\color{blue} $R \leftarrow \mathit{fit\_regressors}(G, X)$; \tcp*[f]{\footnotesize \texttt{learn regressors}}}\\
	\ForEach{$i \in [1, \dots, N]$}{
	    {\color{blue}$z \leftarrow \textsc{gencda}\mathit{\_sampling}(x, G, I, R)$;\tcp*[f]{\footnotesize \texttt{ causal permutations}}}\\
	    $Z \leftarrow Z \cup \{z\}$;\tcp*[f]{\footnotesize \texttt{add synthetic instance}}\\
	    $W \leftarrow W \cup \{exp(\frac{-\pi(x,z)^2}{\sigma^2})\}$;\tcp*[f]{\footnotesize \texttt{add weights}}
    }
    $e \leftarrow \mathit{solve\_Lasso}(Z, b_p(Z), W, k )$;\tcp*[f]{\footnotesize \texttt{get coefficients}}\\
    \Return{$e$}; 
\end{algorithm2e}

The pseudo-code of \calime{} is reported on Algorithm~\ref{alg:calime}, the main differences with \lime{} are highlighted in blue\footnote{We highlight that the \gencda{} algorithm does not explicitly appear in \calime{} pseudo-code as it is decomposed among lines $[2-4]$ and $6$.}.
First, \calime{} runs on $X$ the CD algorithm \ncda{} and extracts the DAG $G$ that describes the causal structure of $X$ (line 2).
After that, for each \textit{root} variable $j$ with respect to $G$, i.e., such that $\mathit{pa}(j) = \emptyset$, \calime{} learns the best distribution that fits $X^{(j)}$ and produces a set of distribution generators $I$ (line 3).
For each \textit{dependent} variable $j$ in $G$, i.e., $\mathit{pa}(j) \neq \emptyset$, \calime{} trains a regressor using as features $X^{\mathit{pa}(j)}$ and as target $X^{(j)}$, and produces a set of regressor generators $R$ (line 4).
For each instance to generate, it runs \gencda{} locally on $x$.
Then, \calime{} takes as input the instance $x$, the DAG $G$, the set of distribution generators for root variables $I$, and the set of regressor generators for dependent variables $R$ (line 6).
The $\textsc{gencda}\mathit{\_sampling}$ randomly selects the features $\{j_1, \dots, j_q\}$ to change among the root ones\footnote{
    The number of features to change $q$ is randomly selected in $[2, q]$.}. 
For each root feature, 
the corresponding data generator $I_j$ generates a synthetic value $z_j$.
After all the root features have been synthetically generated, \calime{} checks the causal relationships modeled in $G$. 
For all dependent variables $j$ such that $\mathit{pa}(j)$ contains a variable that has been synthetically generated, the regressor generator $R_j$ responsible for predicting the value of $j$ is applied on $z^{\mathit{pa}(j)}$ to generate the updated value for feature $j$ by respecting the causal relationship captured in $G$.

To clarify how \calime{} works, we recall the toy example presented at the beginning of this section.
Through \ncda{}, \calime{} discovers a DAG $G$ indicating the causal relation $\text{age} \rightarrow \text{edu\_level}$.
Therefore, it learns the best distributions to model \textit{age}, \textit{income}, and \textit{work\_hours}.
After that it learns a regressor $R_{\text{edu\_level}}$ on $\langle X^{(\text{age})}, X^{(\text{edu\_level})}\rangle$.
Let consider the instance to explain $x = \{(\text{age}, 34), (\text{edu\_level}, 6.5),$ $(\text{income}, 1000), (\text{work\_hours}, 35) \}$.
When a synthetic instance $z$ has to be generated, then \emph{(i)} \textit{education level} can not be changed if \textit{age} is not changed, and \emph{(ii)} when \textit{age} is also changed \textit{education level} must be changed according to $R_{\text{edu\_level}}$.
For instance, if $z_{\text{age}} = 32$ then we can have $z_{\text{edu\_level}} = R_{\text{edu\_level}} = \text{phd-8}$.
This way, the regressor will consider only synthetic customers with a higher \textit{age} when the \textit{education level} is higher. 


%
\section{Experiments}
\label{sec:experiments}
We report here the experiments carried out to validate \calime{}\footnote{
    Python code and datasets are available at \url{https://github.com/marti5ini/CALIME}.
}. 
First, we illustrate the datasets used, the classifiers, and the experimental setup.
Then, we present the evaluation measures adopted. 
Lastly, we demonstrate that our proposal outperforms \lime{} in terms of fidelity, plausibility, and stability. 

\subsection{Datasets and Classifiers}
We experimented with \calime{} on multiple datasets from UCI Repository\footnote{\url{https://archive.ics.uci.edu/ml/datasets/}},
namely: \banknote{}, \magic{}, \wdbc{}, \wine{} and \statlog{}
which belong to diverse yet critical real-world applications.
Table~\ref{tab:datasets} (left) summarizes each dataset. 
The \banknote{} dataset is a well-known benchmark data set for binary classification problems. 
The goal is to predict whether a banknote is authentic or fake based on the measured characteristics of digital images of each banknote. 
The \statlog{} dataset is an image segmentation dataset where the instances were drawn randomly from $7$ outdoor images. 
Each instance is a $3$x$3$ region, and the images were hand-segmented to create a classification for every pixel. 
\wine{} represents wines of different qualities with respect to physicochemical tests from red variants of the Portuguese Vinho Verde wine. 
The \wdbc{} dataset is formed by features computed from a digitized image of a fine needle aspirate of a breast mass. 
They describe the characteristics of the cell nuclei found in the image. 
The \magic{} dataset contains data to simulate high-energy gamma particles in a ground-based atmospheric Cherenkov telescope. 

We would like to highlight that, due to the nature of \gencda{}, all these datasets are tabular datasets with instances represented as continuous features. 
As part of our future research direction, 
we plan to extend \calime{} with a different CD approach capable of handling also categorical and/or discrete variables.

We split each dataset into three partitions: 
$X_{b}$, is the set of records to train a black-box model; 
$X_{c}$, is the set of records reserved for discovering the causal relationships; 
$X_{t}$, is the partition that contains the record to explain. 
%
We trained the following black-box models: Random Forest (\texttt{RF}), and Neural Network (\texttt{NN}) as implemented by \textit{scikit-learn}\footnote{Black-boxes: \url{https://scikit-learn.org/}.}.
For each black-box and dataset, we performed a random search for the best parameter setting\footnote{Detailed information regarding the parameters used can be found in the repository.}.
Classification accuracy and partitioning sizes are shown in Table~\ref{tab:datasets}-(right).


\begin{table}[t!]
    \caption{Dataset statistics and classifiers accuracy. We present the number of records ($n$), the number of features ($m$), the number of labels that the class can assume ($l$), and the number of training records for black-box ($X_b$), the number of explanations records ($X_t$), and the number of records to evaluate the causal relations founded ($X_c$). Additionally, we report Random Forests ($RF$) and Neural Networks ($NN$) accuracy.}    
    \label{tab:datasets}
    \centering
    \resizebox{0.85\textwidth}{!}{
    \begin{tabular}{>{\centering\arraybackslash}p{2cm}||>{\centering\arraybackslash}p{1cm}>{\centering\arraybackslash}p{1cm}>{\centering\arraybackslash}p{1cm}|>{\centering\arraybackslash}p{1cm}>{\centering\arraybackslash}p{1cm}>{\centering\arraybackslash}p{1cm}||>{\centering\arraybackslash}p{1cm}>{\centering\arraybackslash}p{1cm}}
      & $n$ & $m$ & $l$ & $|X_{b}|$ & $|X_{c}|$ & $|X_{t}|$ & $RF$ & $NN$ \\
    \midrule
    \banknote{} & 1,372 & 4  & 2 & 890 & 382 & 100 & .99 & 1.0 \\
    \statlog{}  & 2,310 & 19 & 7 & 1,547 & 663 & 100 & .98 & .96 \\
    \magic{}    & 19,020 & 11 & 2 & 13,244 & 5,676 & 100 & .92 & .85\\
    \wdbc{}     & 569 & 30 & 2 & 328 & 141 & 100 & .95 & .92\\
    \wine{}    & 1,159 & 11 & 6 & 358 & 203 & 154 & .82 & .70\\
    \end{tabular}
    }
\end{table}

\subsection{Comparison with Related Works}
We evaluate the effectiveness of the \calime{} framework through a comparative analysis, contrasting it to several state-of-the-art proposals designed to outperform the limitations of \lime{}.
We specifically select two baseline methods: \textsc{f-lime}~\cite{saito2020improving}, which employs a Conditional Tabular Generative Adversarial Network to create more realistic synthetic data for model explanations, and \textsc{d-lime}~\cite{zafar2021deterministic}, a Deterministic Local Interpretable Model-Agnostic Explanations approach that replaces random perturbations with hierarchical clustering to group the data.

The rationale for selecting these specific methods is grounded in several key factors: i) similar to \calime{}, they innovate by altering the neighborhood generation mechanism while preserving the core algorithmic structure; ii) they have been developed relatively recently; iii) the presence of accessible source code enhances reproducibility. For their implementation, we used the versions available in the respective repository\footnote{The \textsc{f-lime} code can be accessed at \url{https://github.com/seansaito/Faster-LIME}, while \textsc{d-lime} at \url{https://github.com/rehmanzafar/dlime_experiments}}. 
The parameter settings followed the original paper.

\subsection{Evaluation Measures}
We evaluate the quality of the explanations returned to three criteria, i.e., \textit{fidelity}, \textit{plausibility}, and \textit{stability} of the explanations.

\smallskip
\textbf{Fidelity.}
One of the metrics most widely used in XAI to evaluate how good an explainer is at mimicking the black-box decisions is the \textit{fidelity}~\cite{guidotti2019survey}. 
There are different specializations of fidelity, depending on the type of explanator under analysis~\cite{guidotti2019survey}.
In our setting, we define the fidelity in terms of the coefficient of determination $R^2$:
$$R_x^2 = 1 - \frac{\sum_{i=1}^{N}(b(z_i) - r(z_i))^2}{\sum_{i=1}^{N}(b(z_i) - \hat{y})^2}\;\;\;\text{with}\;\;\; \bar{y}=\frac{1}{N}\sum_{i=1}^{N}b(z_i)$$
where $z_i \in Z$ is the synthetic neighborhood 
for a certain instance $x$, and $r$ is the linear regressor with Lasso regularization trained on $Z$.
$R^2$ ranges in $[0, 1]$ and a value of $1$ indicates that the regression predictions perfectly fit the data.

\smallskip
\textbf{Plausibility.}
We evaluate the plausibility of the explanations regarding the goodness of the synthetic datasets locally generated by \lime{} and \calime{} by using the following metrics based on distance, outlierness, statistics, likelihood, and detection. 
In~\cite{patki2016synthetic} is presented a set of functionalities that facilitates evaluating the quality of synthetic datasets. 
Our experiments exploit this framework to compare the synthetic data of the neighborhood $Z$ with $X_b$.

\textit{Average Minimum Distance Metric.}
Given the local neighborhood $Z$ generated around instance $x$, 
a synthetic instance $z_i \in Z$ is plausible if it is not too much different from the most similar instance in the reference dataset $X$.

Hence, for a given explained instance $x$, we calculate the plausibility in terms of Average Minimum Distance ($AMD$): 
$$AMD_x = \frac{1}{N}\sum_{i=1}^N d(z_i, \bar{x})\;\;\;\text{with}\;\;\; \bar{x}=\arg\min_{x' \in X / \{x\}}d(z_i, x')$$
where the lower the $AMD$, the more plausible the instances in $Z$, the more reliable the explanation.

\textit{Outlier Detection Metrics.}
We also evaluate the plausibility of outliers in the synthetic neighborhood $Z$.
The fewer outliers are in $Z$ to $X$, the more reliable the explanation is.
In particular, we estimate the number of outliers in $Z$ by employing three outlier detection techniques\footnote{LOF and IF: \url{https://scikit-learn.org/stable/modules/generated/sklearn.neighbors}; ABOD: \url{https://pyod.readthedocs.io}}: Local Outlier Factor (LOF), 
Angle-Based Outlier Detection (ABOD), 
and Isolation Forest (IF)~\cite{chandola2009anomaly}.
These three approaches return a value in $[0, N]$, indicating the number of outliers identified by the method.
We report the Average Outlier Score ($AOS$) that combines the normalized scores of these indicators.

\textit{Statistical Metrics.}
To compare $X$ with $Z$, we exploit the statistical measures outlined in SDV~\cite{patki2016synthetic}.
Specifically, we opted for the Gaussian Mixture Log Likelihood, (\texttt{GMLogLikelihood}), 
Inverted Kolmogorov-Smirnov D statistic, (\texttt{KSTest}), 
and Continuous Kullback-Leibler Divergence (\texttt{ContinuousKLDivergence}). \\
\texttt{GMLogLikelihood} fits multiple Gaussian Mixture models to the real data using different numbers of components and returns the average log-likelihood given to the synthetic data.
\texttt{KSTest} performs the two-sample Kolmogorov–Smirnov test using the empirical Cumulative Distribution Function (CDF). The output for each column is one minus the KS Test D statistic, which indicates the maximum distance between the expected CDF and the observed CDF values.
\texttt{ContinuousKLDivergence} is an information-based measure of disparity among probability distributions. 
The SDV framework approximates the KL divergence by binning the continuous values into categorical values and then computing the relative entropy. Afterward, normalize the value using $1/(1 + D_{KL})$. 
The metric is computed as the average across all the column pairs.
We report the Average Statistical Metric ($ASM$) that aggregates these three indicators.

\begin{figure}[t]
  \centering
  \begin{subfigure}[b]{0.55\textwidth}
    \includegraphics[width=\textwidth]{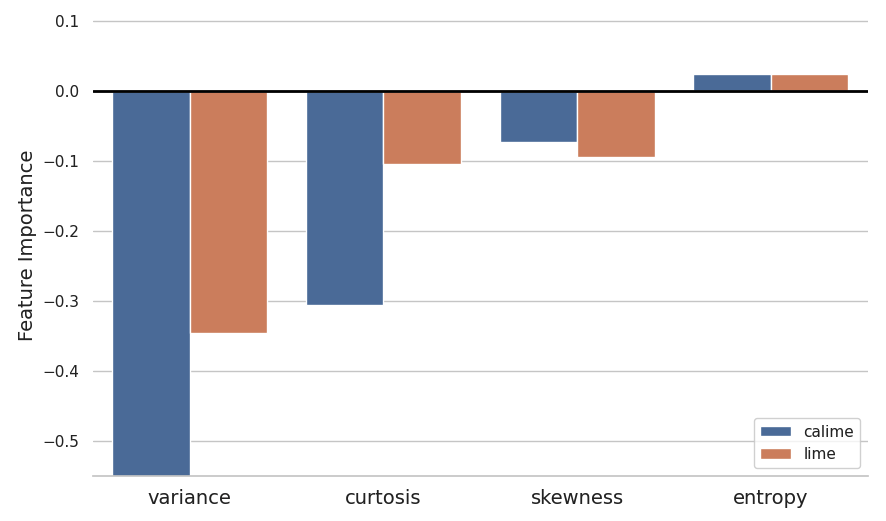}
  \end{subfigure}%
  \hfill
  \begin{subfigure}[b]{0.4\textwidth}
    \includegraphics[width=\textwidth]{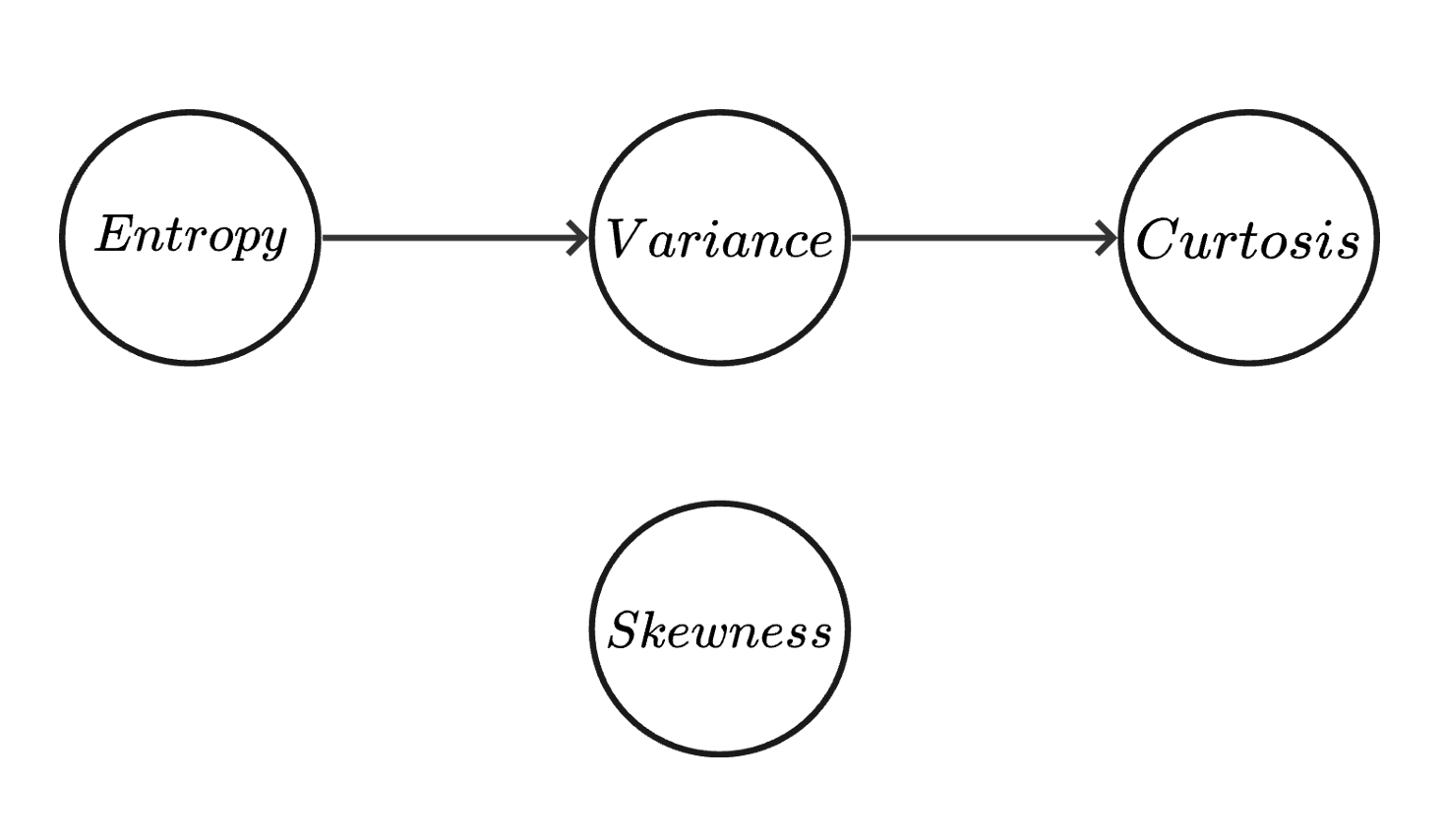}
  \end{subfigure}
  \caption{(Left). Feature importance computed by \calime{} and \lime{}. (Right). Causal graph of \banknote{} inferred by a causal discovery method~\cite{hoyer2008nonlinear}.}
  \label{fig:graph_banknote}
\end{figure}

\textit{Detection Metrics.} 
A way to test the plausibility of a synthetic dataset is to use a classification approach to assess how much the real data differs from synthetic ones~\cite{patki2016synthetic}. 
The idea is to shuffle the real and synthetic data together, label them with flags indicating whether a specific record is real or synthetic, and cross-validate an ML classification model that tries to predict this flag~\cite{patki2016synthetic}. 
The output of the metric is one minus the average Area Under the Receiver Operating Curve across all the cross-validation splits.
We employ a Logistic detector and SVM as ``discriminators'' and report the Average Detection Metric ($ADM$).

\smallskip
\textbf{Stability.} 
To gain the users' trust, explanation methods must guarantee stability across different explanations~\cite{guidotti2019stability}.
Indeed, the stability of explanations is a fundamental requirement for a trustworthy approach~\cite{gosiewska2019not}.
Let $E=\{e_1, \dots, e_n\}$ be the set of explanations for the instances $X=\{x_1, \dots, x_n\}$.
In line with~\cite{guidotti2019black}, we asses the \textit{stability} through the local Lipschitz estimation~\cite{melis2018towards}: 
$$LLE_x = \underset{x_i \in \mathcal{N}^k_x}{avg} \frac{\lVert e_i - e \rVert_2}{ \lVert x_i - x \rVert_2}$$
where $x$ is the instance to explain and $\mathcal{N}^k_x \subset X$ is the $k$-Nearest Neighborhood of $x$ with the $k$ neighbors selected from $X_t$.
The lower the $LLE$, the higher the stability.
In the plots illustrating the stability through $LLE$, besides the average value, we also report the minimum and maximum, using them as contingency bands.
According to the literature~\cite{melis2018towards}, we used $k=5$.

\begin{figure}[t]
    \centering
    \includegraphics[width=.33\linewidth]{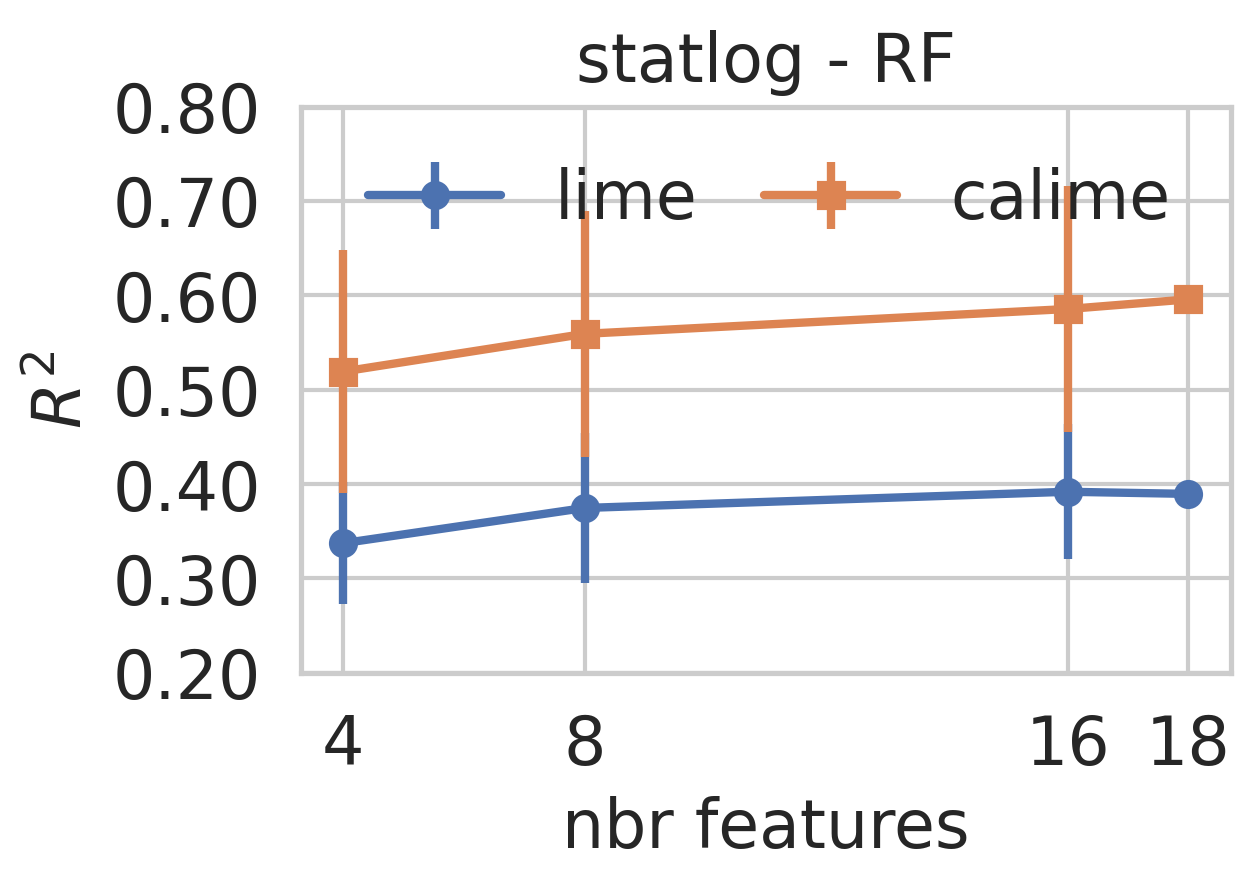}%
    \includegraphics[width=.33\linewidth]{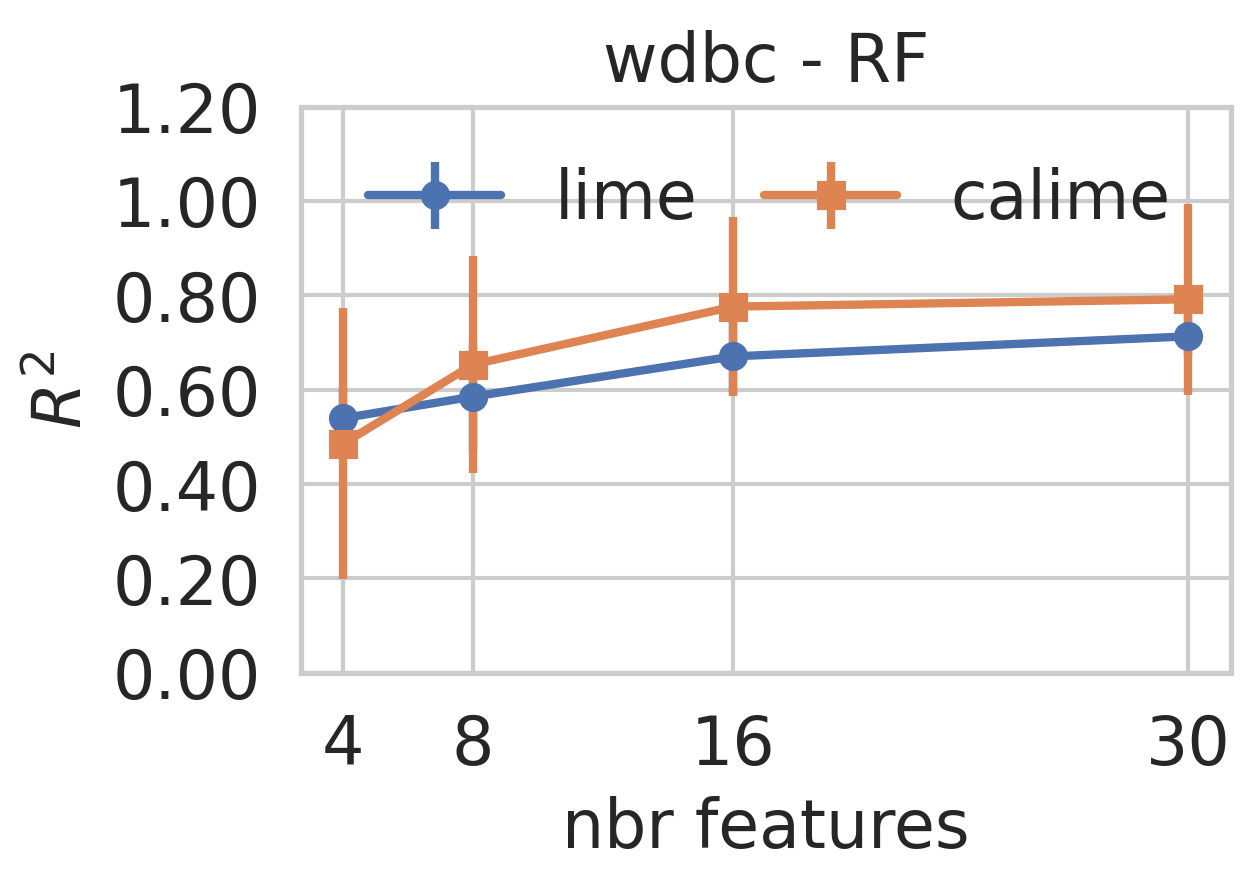}%
    \includegraphics[width=.33\linewidth]{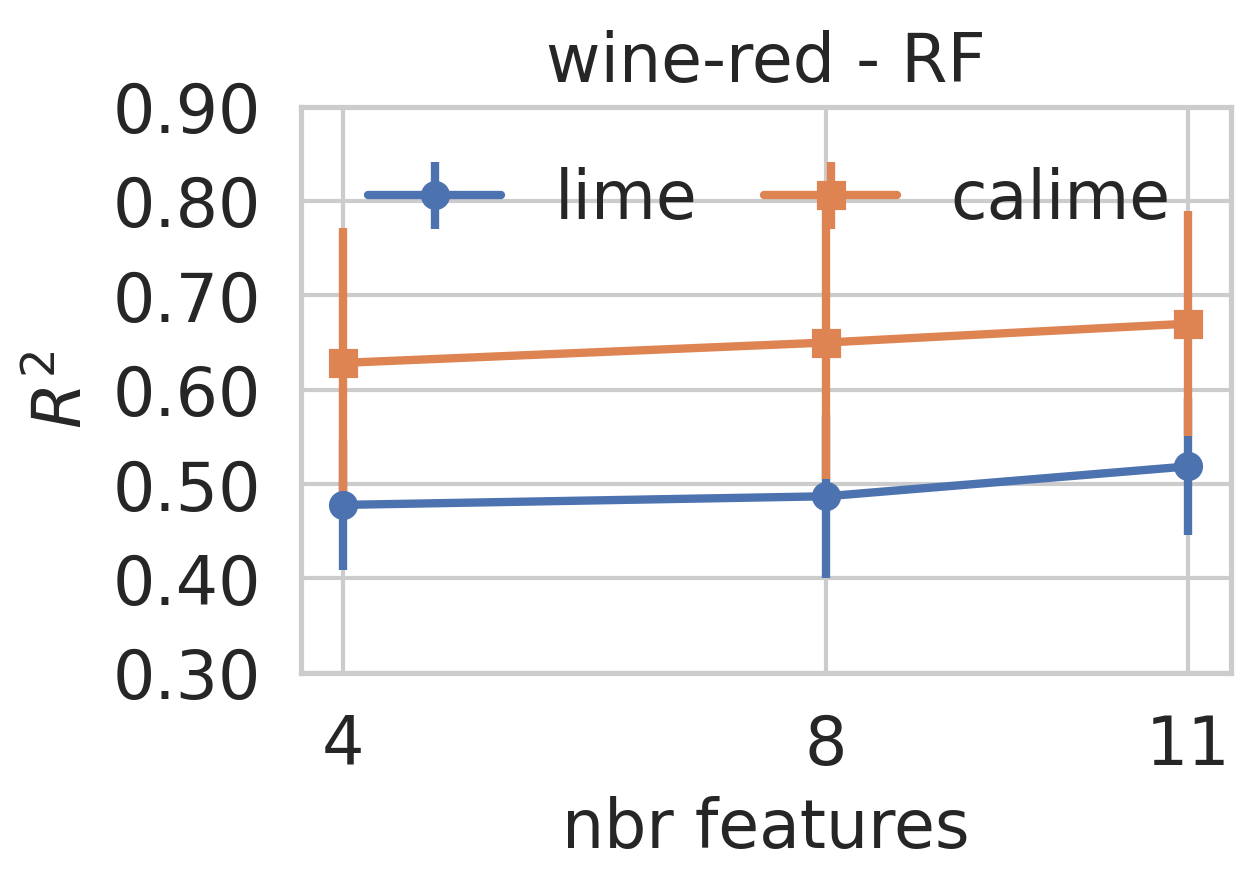}
    \includegraphics[width=.33\linewidth]{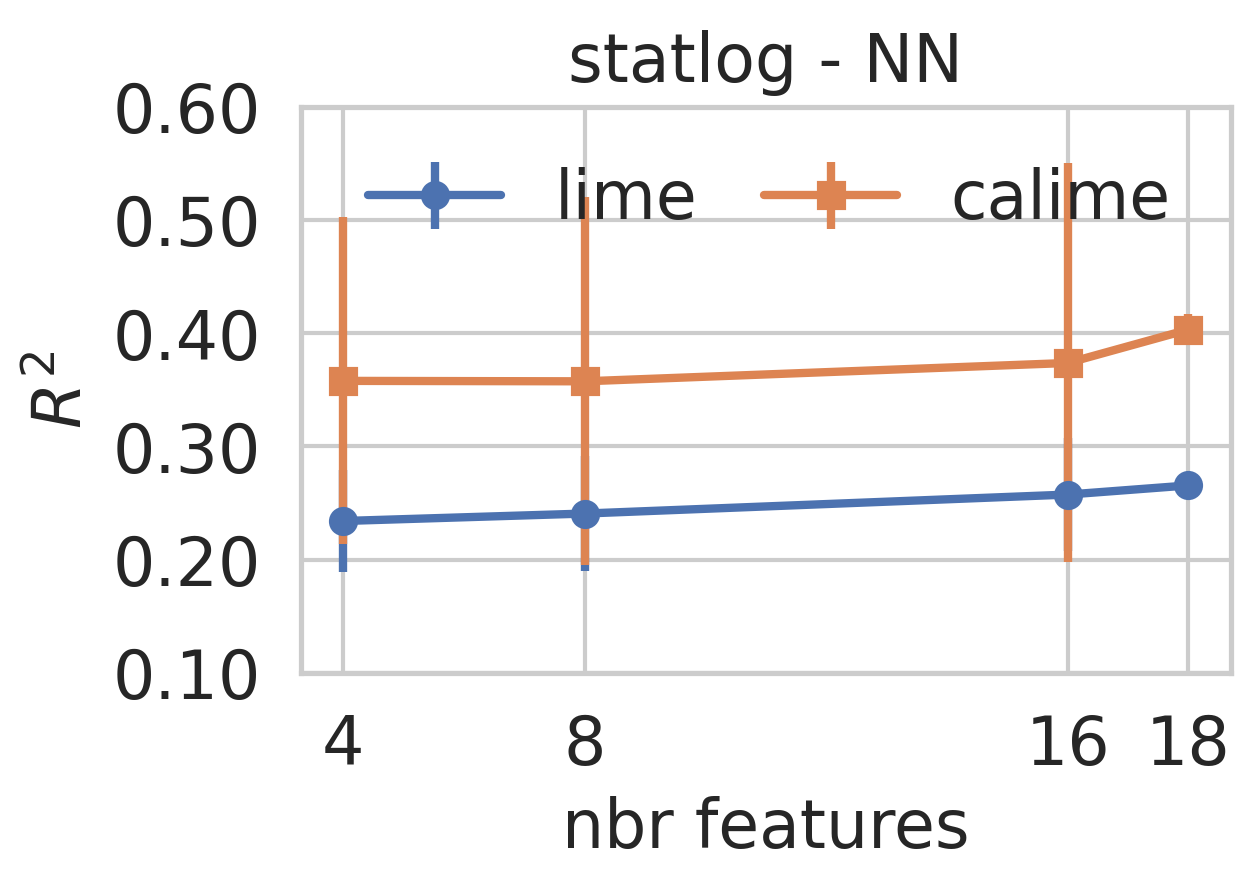}%
    \includegraphics[width=.33\linewidth]{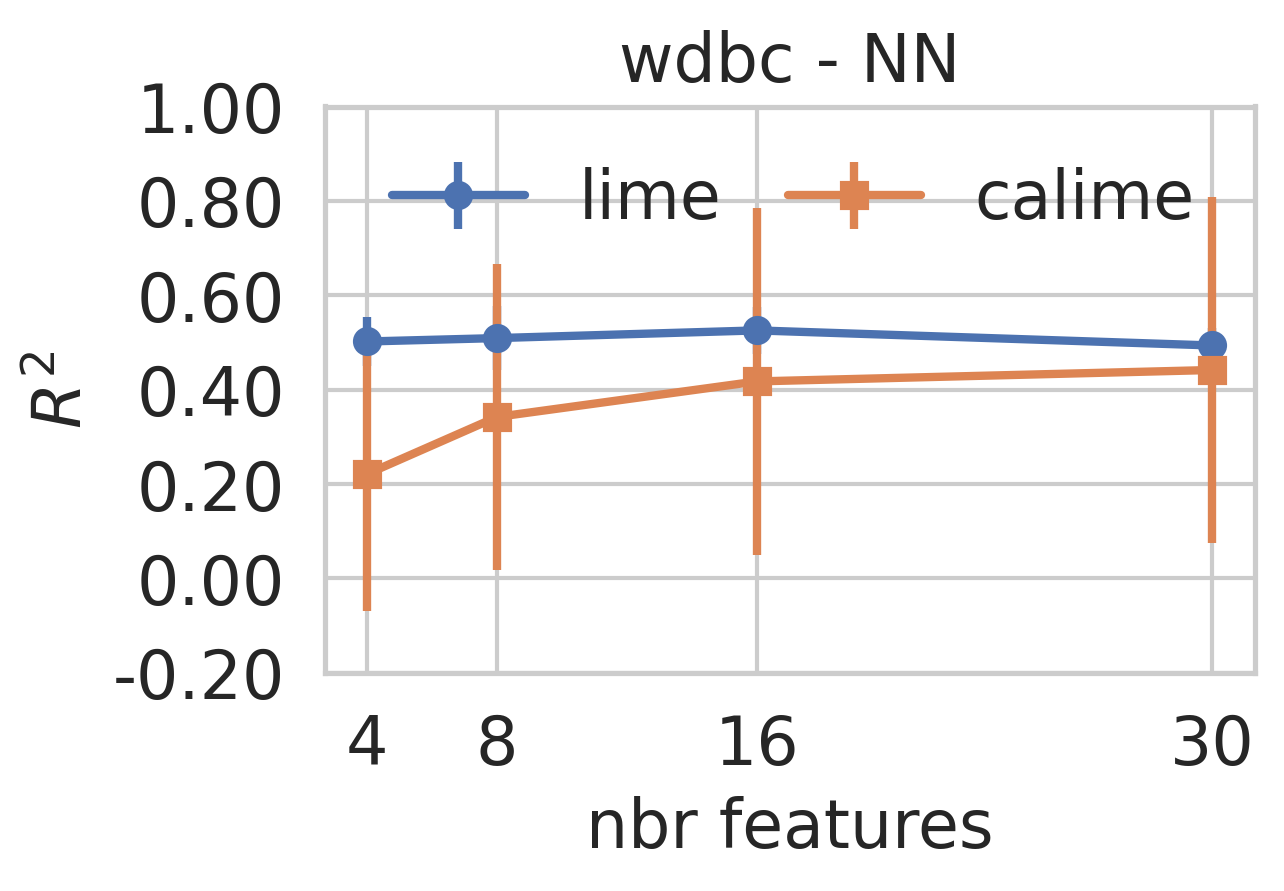}%
    \includegraphics[width=.33\linewidth]{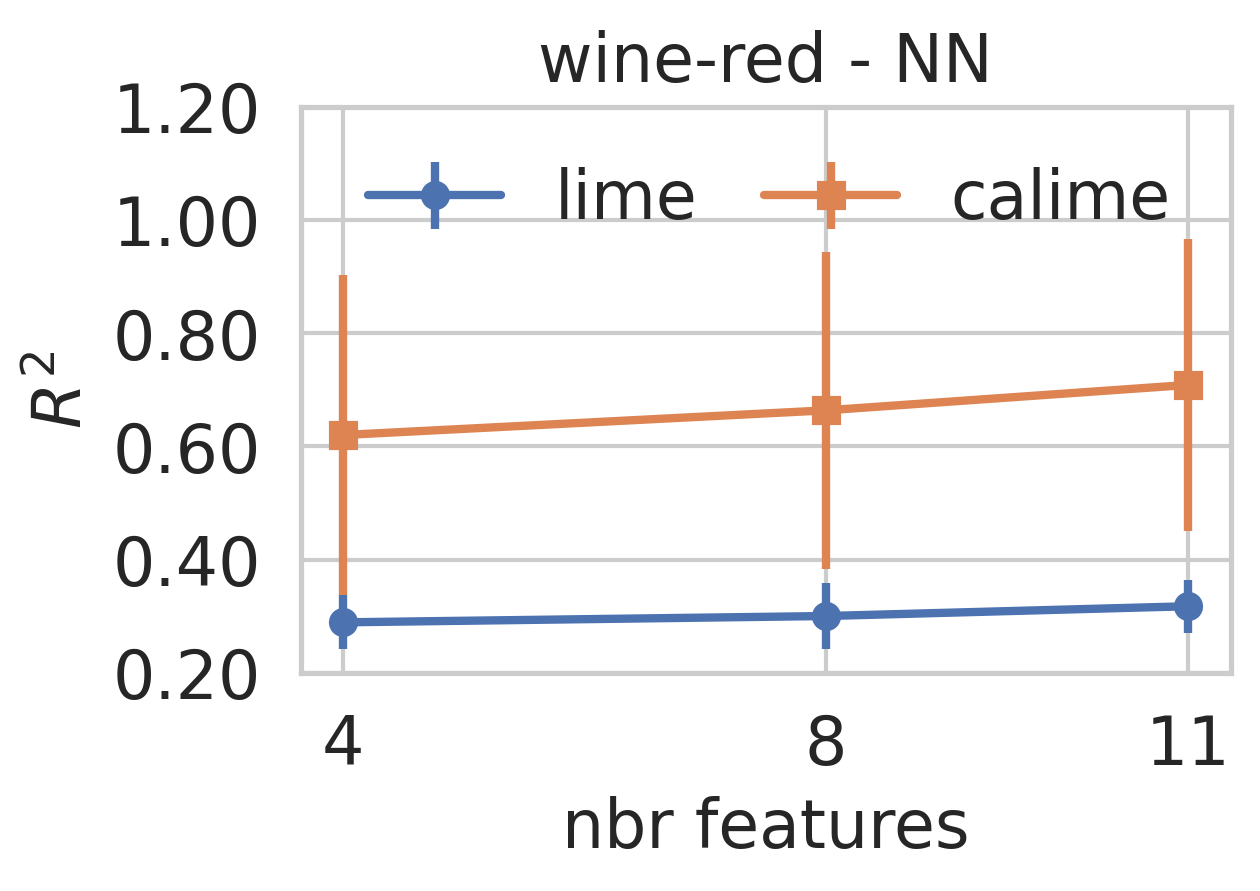}
    \caption{Fidelity as $\boldsymbol{R^2}$ varying the number of features for \statlog{}, \wdbc{} and \wine{}. Markers represent the mean values, while vertical bars are the standard deviations.}
    \label{fig:fidelity}
\end{figure}

\subsection{Results}
We propose a qualitative comparison between \calime{} and \lime{} by examining the explanations generated by both approaches for a specific instance within the \banknote{} dataset (Figure \ref{fig:graph_banknote} - Left). 
It is important to note that \calime{} is based on the causal framework illustrated in Figure \ref{fig:graph_banknote} on the right discovered using a causal discovery method~\cite{hoyer2008nonlinear}, where \textit{entropy} causes \textit{variance}, which in turn influences \textit{curtosis}\footnote{To avoid misinterpretation of these variables, it is important to note that their names are not intended to refer to their mathematical meanings. }.  
This sequence mirrors the actual physical process whereby the textural features of a banknote determine its classification as genuine or counterfeit. 
We notice that the feature importance scores for \textit{variance} and \textit{curtosis} obtained through \calime{} are significantly higher compared to those attained using \lime{}. This is due to the fact that the neighborhood of the explained record is generated by adhering to the causal relationships outlined in the causal graph.
Conversely, \lime{}  may not adequately capture this causal sequence, potentially undervaluing \textit{variance} due to its dependence on correlation, which can be confounded by spurious relationships in the data.
Hence, \calime{} offers a more contextually relevant explanation by incorporating the causal relationships intrinsic to banknote verification processes. It ensures that model explanations are also meaningful, reflecting the dynamics involved in the real-world task.

In Figure~\ref{fig:fidelity} we show the fidelity as $R^2$ of \lime{} and \calime{} for \statlog{}, \wdbc{} and \wine{} varying the number of features $k$. 
Results show that \calime{} provides an improvement to \lime{} except \wdbc{} for which the performance of \calime{} is slightly worse than \lime{}. 
More in detail, for $k=30$, \lime{} is more faithful for the NN classifier trained on \wdbc{} since it achieves $.44$ in contrast to $.41$ for \calime{}. 
For the standard deviation, we notice that \calime{} is more stable than \lime{}.
To summarize, \calime{} provides explanations more trustworthy than those returned by \lime{} at the cost of a slightly larger variability in the fidelity of the local models. 

\begin{figure}[!t]
    \centering
    \includegraphics[width=.45\linewidth]{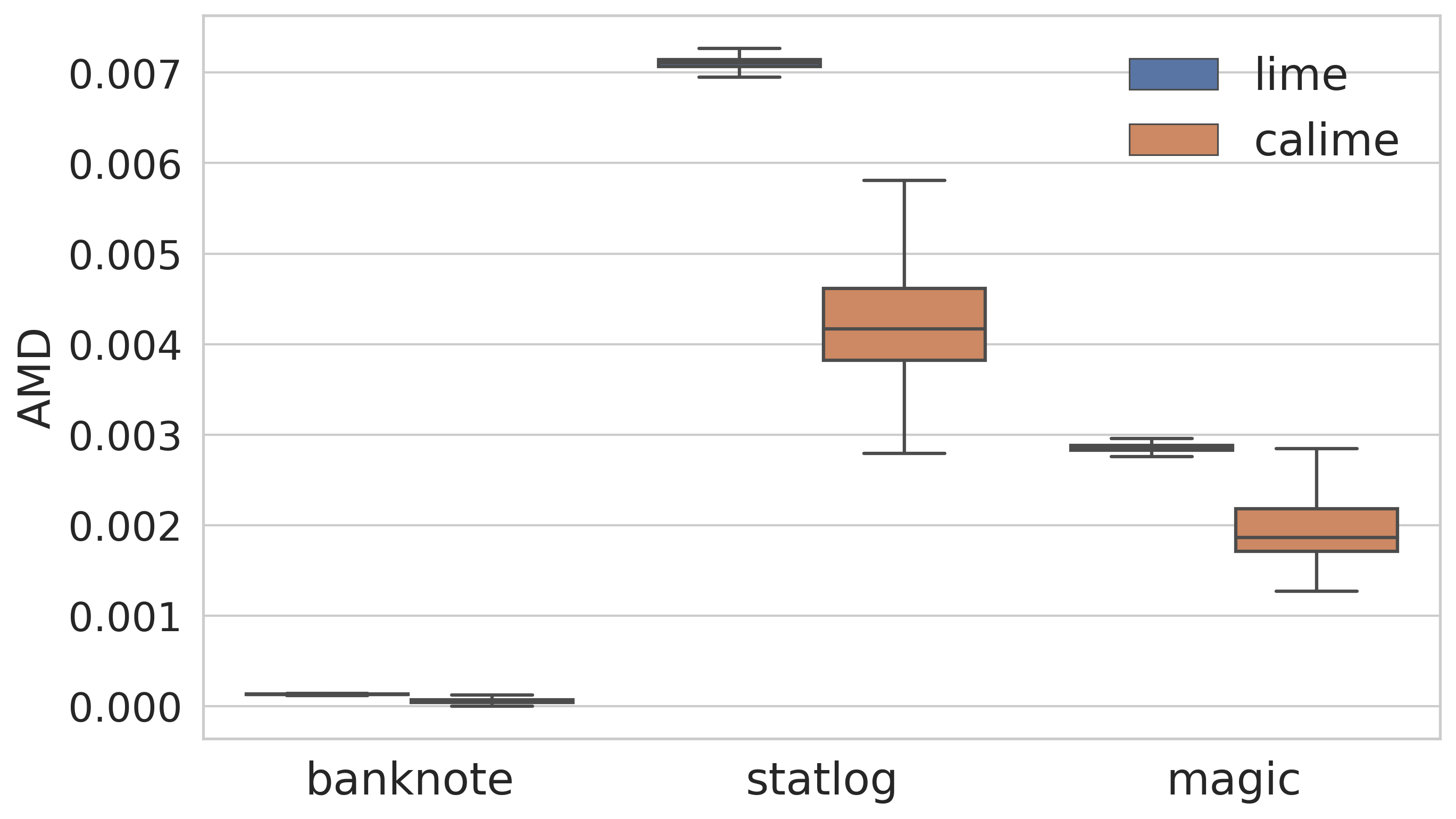}%
    \includegraphics[width=.45\linewidth]{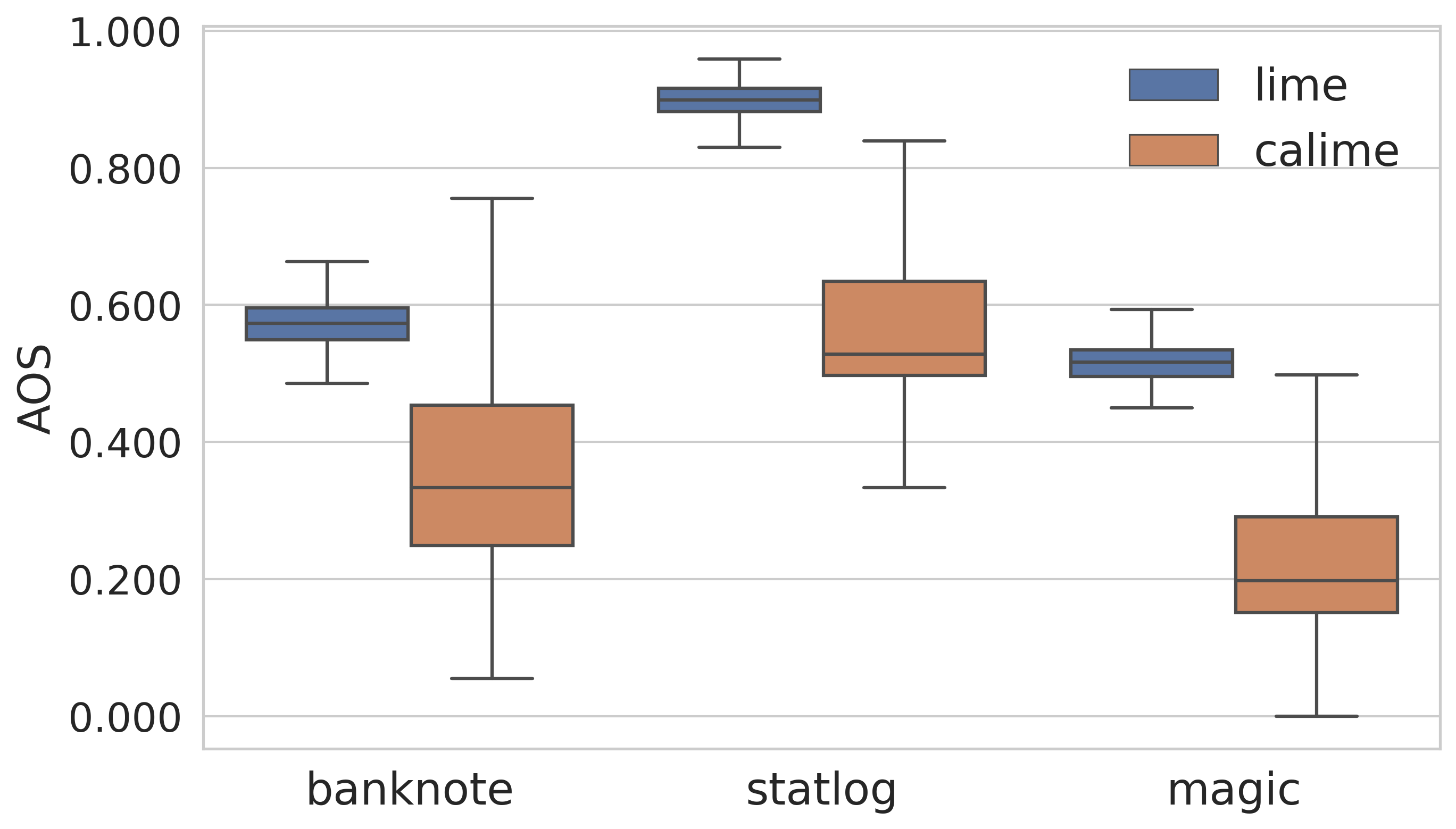}
    \includegraphics[width=.45\linewidth]{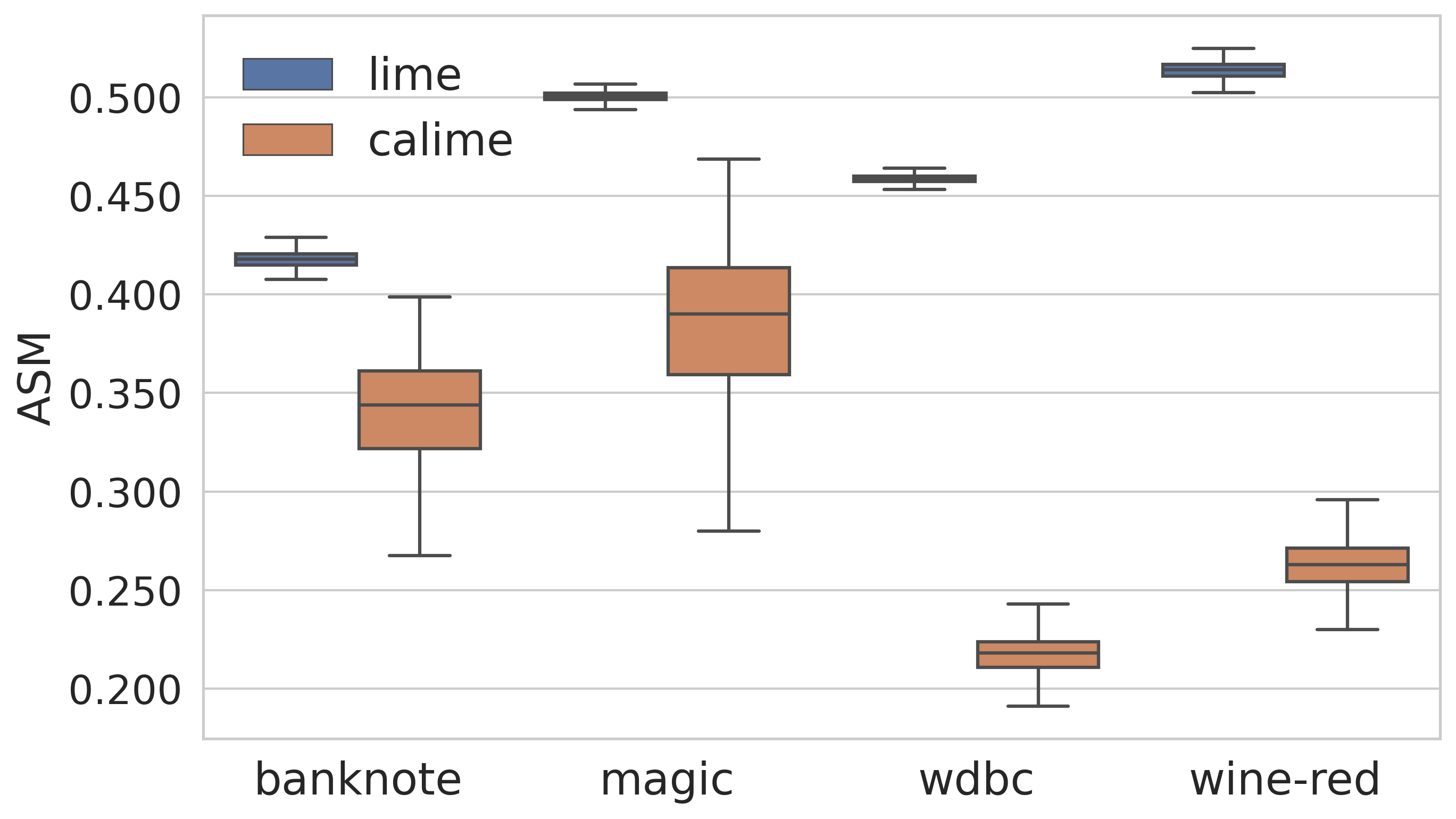}%
    \includegraphics[width=.45\linewidth]{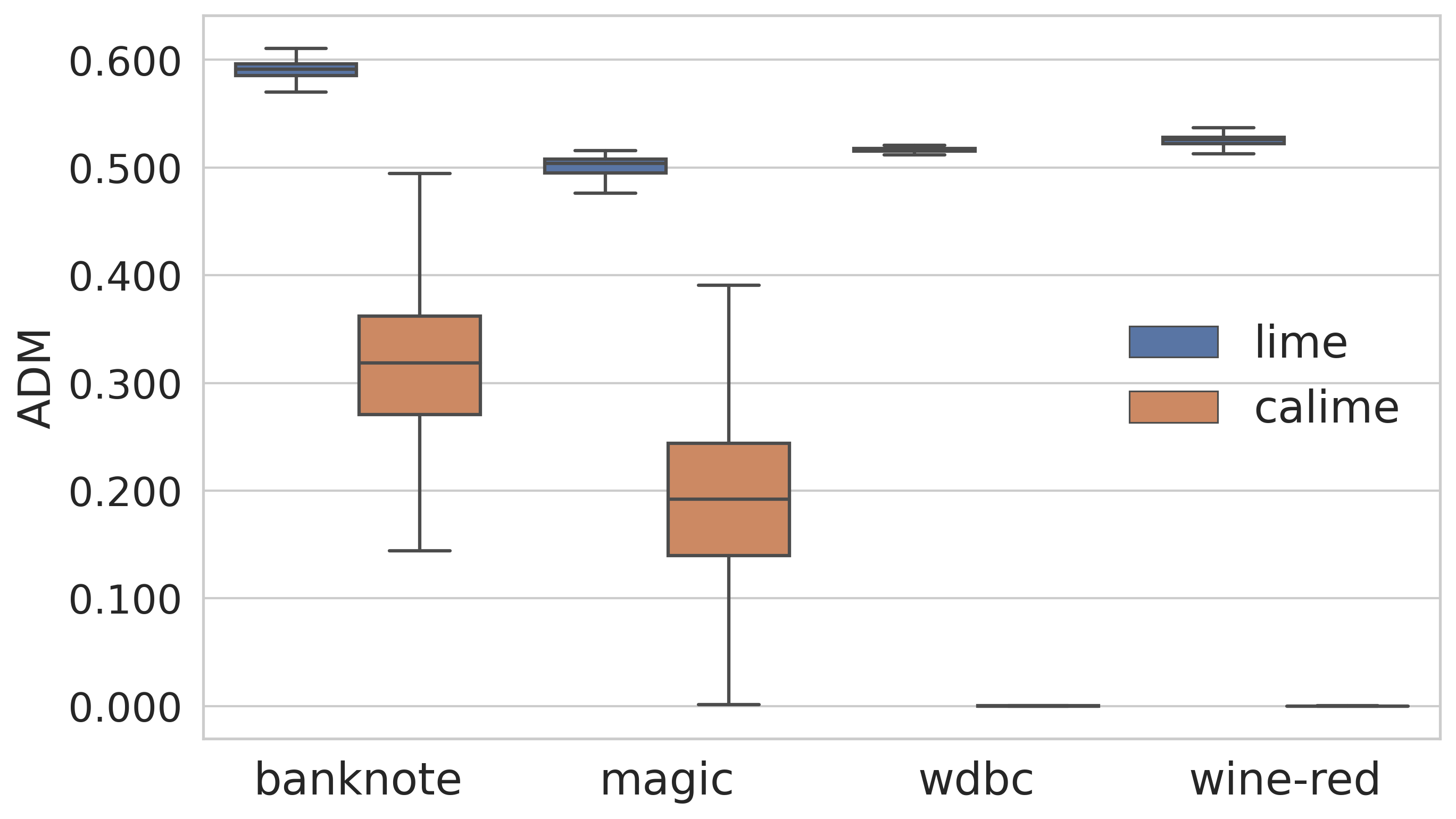}
    \caption{Plausibility errors as $AMD$, $AOD$, $ASM$, and $ADM$ in box plots aggregating the results across the scores obtained with different numbers of features. Best viewed in color.}
    \label{fig:plausibility}
\end{figure}

In Figure~\ref{fig:plausibility}, we illustrate the plausibility of \lime{} and \calime{} through box plots aggregating the results obtained for the different number of features $k$ for the selected datasets considering simultaneously results for the different black-boxes.
We remind the reader that all the scores observed, i.e., $AMD$, $AOD$, $ASM$, and $ADM$, are ``sort of'' errors: lower values indicate higher plausibility scores. 
Analyzing the results, we observe that \calime{} outperforms \lime{} in terms of plausibility for all the metrics analyzed.
The enhancement in \calime{}'s performance can be credited to the acquired knowledge of relationships among the dataset variables, which enables the generation of data points in the local neighborhood of $x$ that closely resemble the original data. 
Concerning the $AMD$ and $AOS$ scores, we observe the results focusing the analysis on \banknote{}, \magic{} and \statlog{}. 
In general, \calime{} consistently produces a closer and more condensed neighborhood. In contrast, the randomly generated samples by \lime{} are further apart and exhibit lower density around the instance being explained.
For the \banknote{} dataset, we notice relatively similar results between the two methods, whereas the most noticeable difference, particularly in terms of $AMD$ and $AOS$, becomes evident when analyzing \statlog{}.
Regarding the $ASM$ and $AMD$ metrics, for which we provide insights concerning \banknote{}, \magic{}, \wdbc{}, and \wine{}, \calime{} offers a more realistic and probable synthetic local neighborhood compared to \lime{}.
Notably, the datasets that show the most significant improvement in \lime{}'s performance are \wdbc{} and \wine{}.
Consistent with observations in fidelity, the greatest plausibility of \calime{} correlates with a more diverse neighborhood, as evidenced by the larger box plots (e.g., for \magic{}).

\begin{figure}[t]
    \centering
    \includegraphics[width=.33\linewidth]{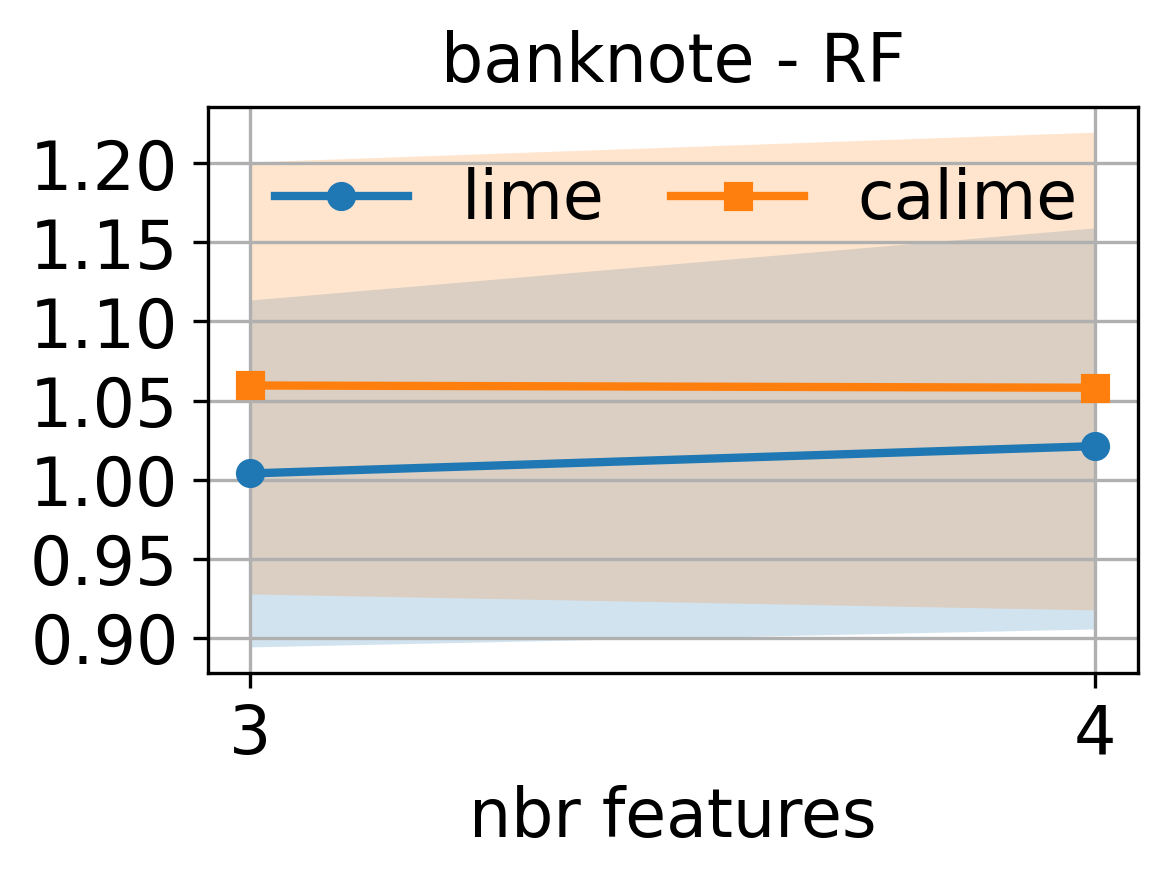}%
    \includegraphics[width=.33\linewidth]{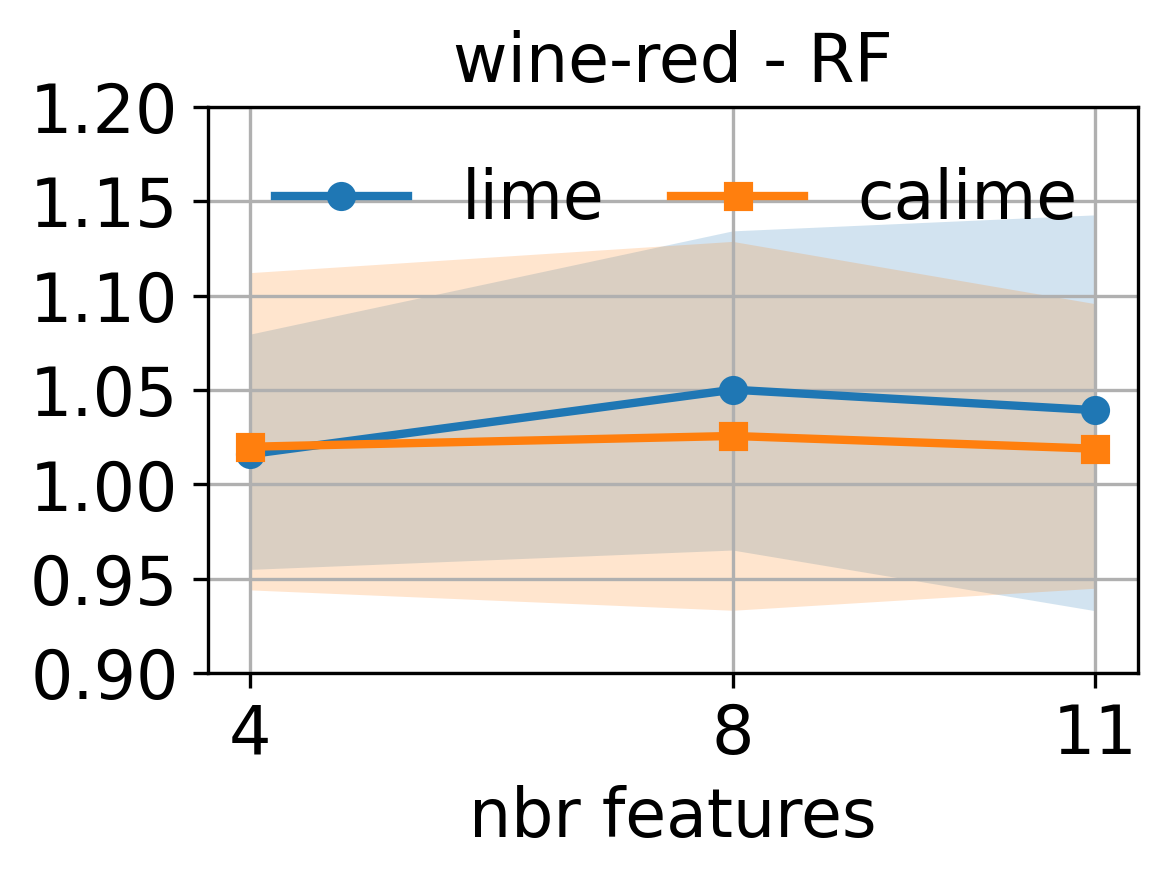}%
    \includegraphics[width=.33\linewidth]{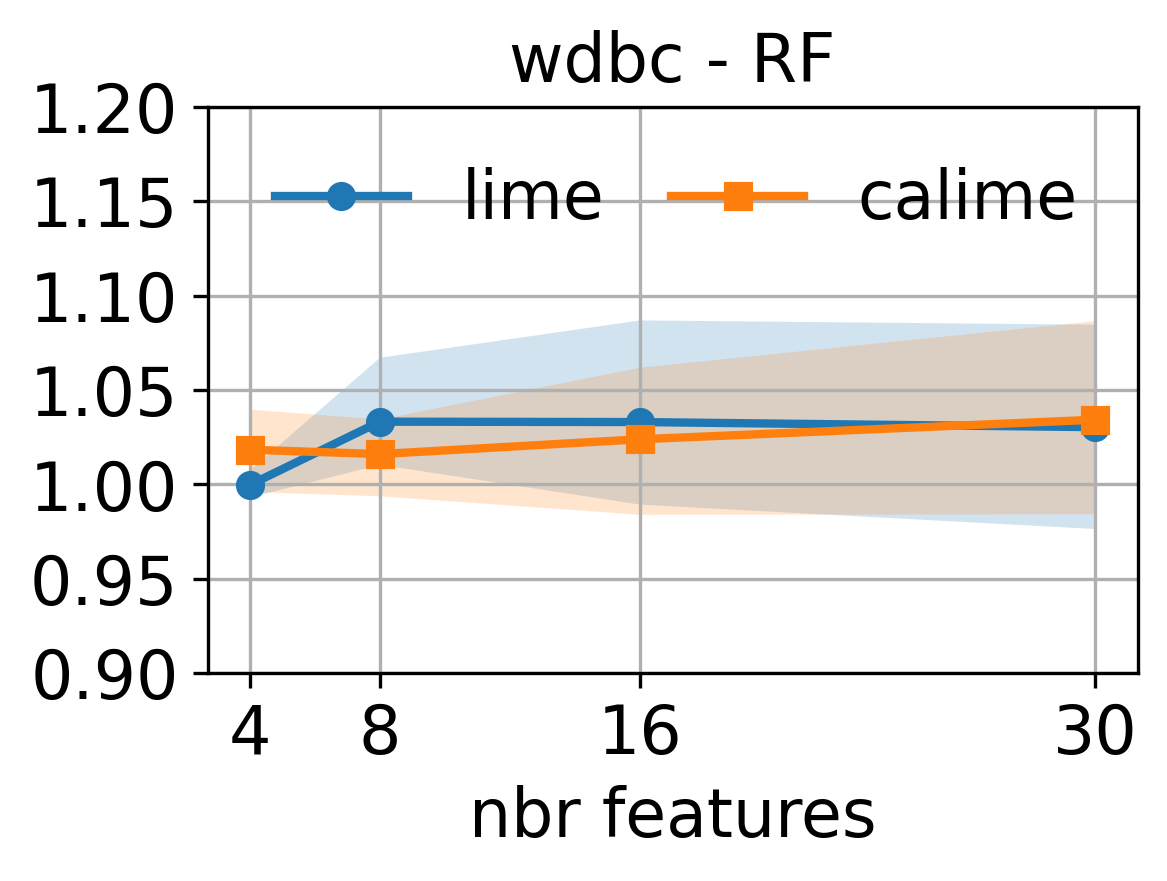}
    \includegraphics[width=.33\linewidth]{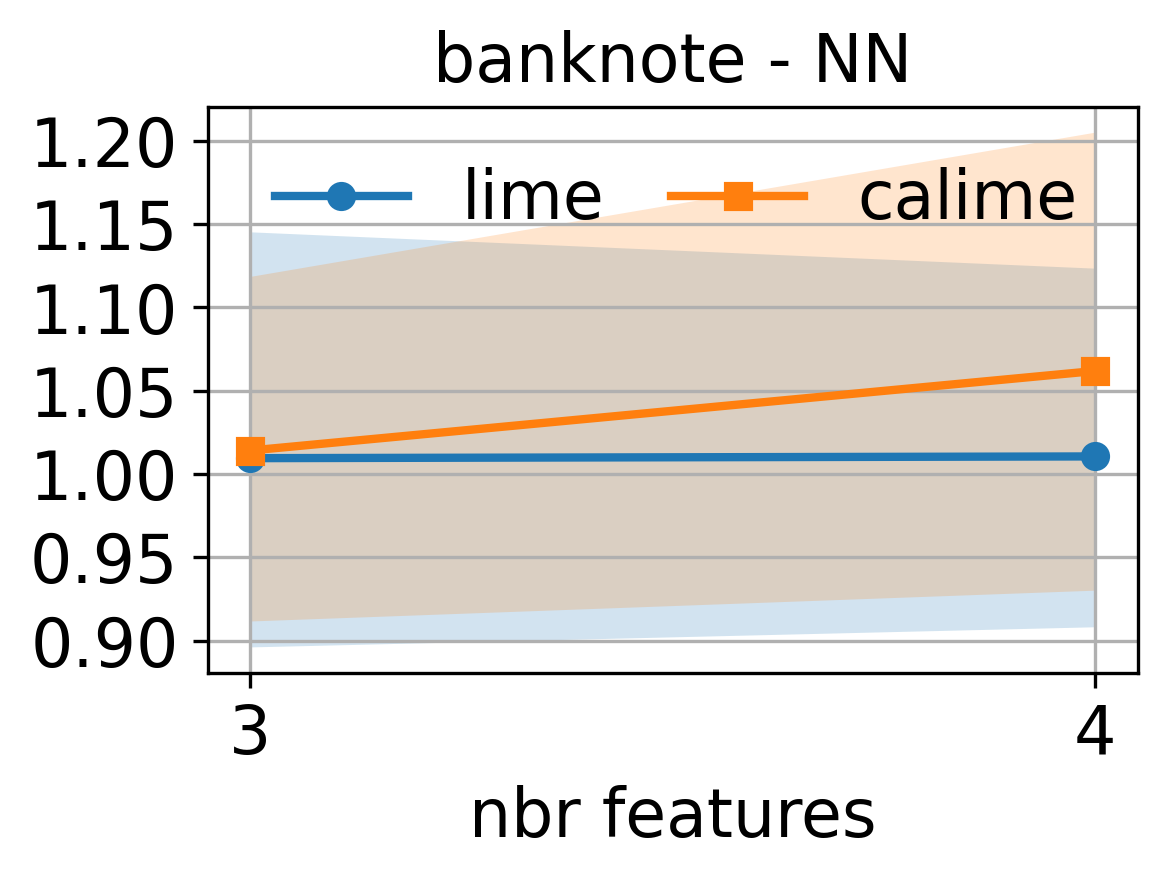}%
    \includegraphics[width=.33\linewidth]{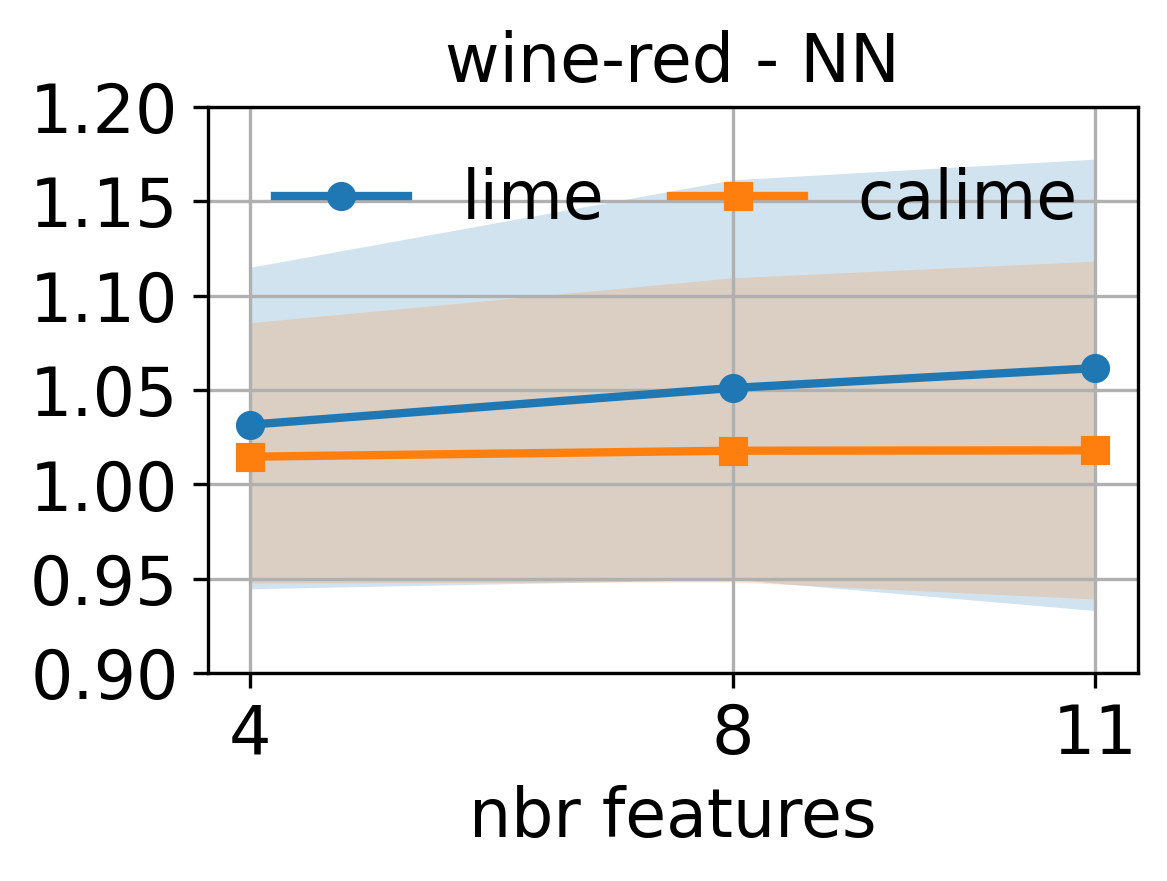}%
    \includegraphics[width=.33\linewidth]{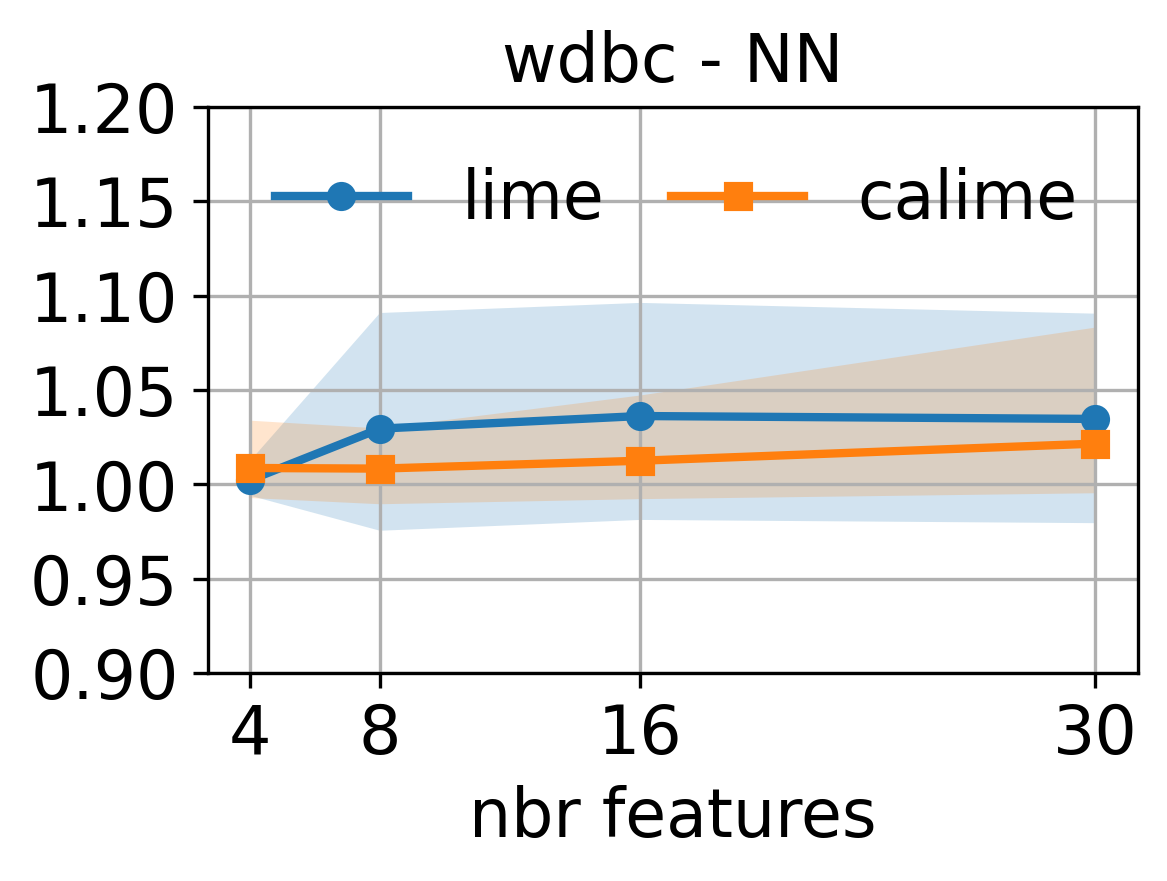}
    \caption{Instability as $\boldsymbol{LLE}$ varying the number of features. Markers represent the mean values, while the contingency area highlights the minimum and maximum values.}
    \label{fig:stability}
\end{figure}

In Figure~\ref{fig:stability}, we illustrate the instability of \lime{} and \calime{} as $LLE$ (the lower, the better) for \banknote{}, \wine{} and \wdbc{} varying the number of features $k$.
The colored areas highlight the minimum and maximum values of $LLE$ obtained by replacing the min and max operators in the previous formula.
We notice that \lime{} is more stable than \calime{} or comparable when the number of features perturbed by the neighborhood generation procedures is small, i.e., $k \leq 4$.
Conversely, \calime{} is more resistant to noise than \lime{} when $k>4$. 

Finally, we show the comparison between \calime, \lime{}, and two other state-of-the-art methods that replace the neighborhood generation in the original algorithm to achieve explanations with higher fidelity and plausibility. In Table \ref{tab:comparison}, we provide the results obtained w.r.t. these metrics. 
Regardless of how the neighborhood is generated, all methods perform better than \lime{}. Our approach excels, especially in terms of $AMD$, $AOS$, and $R^2$ when compared to all other methods. About $ADM$ and $ASM$, the results are nearly comparable between \calime{} and \textsc{f-lime}. This similarity may be attributed to the GAN-based generation used by \textsc{f-lime}. 
In theory, GAN-like methods have the potential to capture possible dependencies. However, as empirically demonstrated in~\cite{cinquini2021gencda}, these relationships are not explicitly represented, and there is no guarantee that they are faithfully followed in the data generation process.

In summary, \calime{} empirically exhibits better performance than \lime{} on the datasets and the black-boxes analyzed.
However, one drawback of \calime{}, at least concerning the current implementation, lies in its slower execution, primarily due to the additional time overhead associated with two specific tasks: \emph{(i)} the extraction of the DAG and \emph{(ii)} the learning of probability distributions for root variables and regressors to approximate dependent variables.
For instance, while \lime{} can generate explanations in less than a second, \calime{} necessitates one order of magnitude more time for completion.

\begin{table}[t!]
\caption{A comparative analysis of Fidelity and Plausibility between \calime{} and baseline methods.}
\label{tab:comparison}
\centering
\resizebox{0.85\textwidth}{!}{
\begin{tabular}{>{\centering\arraybackslash}p{1.5cm} >{\centering\arraybackslash}p{1.8cm} >{\centering\arraybackslash}p{1.5cm} >{\centering\arraybackslash}p{1.5cm} >{\centering\arraybackslash}p{1.5cm} >{\centering\arraybackslash}p{1.5cm} >{\centering\arraybackslash}p{1.5cm}}
Datasets & Explainers & AMD $\downarrow$ & AOS $\downarrow$ & ASM $\downarrow$ & ADM $\downarrow$ & $R^2$ $\uparrow$  \\ 
\midrule
{\multirow{4}{*}{\banknote{}}}
& \lime{} & $.188$ & $459$ & $.409$ & $.583$ & $.657$ \\
& \calime{} & $\textbf{.085}$ & $\textbf{236}$ & $.346$ & $.412$ & $\textbf{.661}$ \\
& \textsc{f-lime} & $.109$ & $428$ & $\textbf{.336}$ & $\textbf{.337}$ & $.643$ \\
& \textsc{d-lime} & $.193 $ & $672$  & $.360$ & $.502$ & $.608$ \\
\hline
{\multirow{4}{*}{\wine{}}} 
& \lime{} & $.234$ & $471$ & $.518$ & $.675$ & $.489$ \\
& \calime{} & $\textbf{.092}$ & $\textbf{380}$ & $.287$ & $.524$ & $\textbf{.683}$
\\
& \textsc{f-lime} & $.213$ & $400$ & $\textbf{.106}$ & $.510$ & $.625$ \\
& \textsc{d-lime} & $.207$ & $293$ & $.173$ & $\textbf{.503}$ & $.611$ \\
\hline
{\multirow{4}{*}{\statlog{}}} 
& \lime{} & $.608$ & $794$ & $.353$ & $.490$ & $.398$ \\
& \calime{} & $\textbf{.487}$ & $\textbf{349}$ & $.265$ & $.418$ & $\textbf{.453}$ \\
& \textsc{f-lime} & $.553$ & $441$ & $\textbf{.216}$ & $\textbf{.375}$ & $.446$ \\
& \textsc{d-lime} & $.610$ & $597$ & $.280$ & $.478$ & $.413$ \\
\hline
{\multirow{4}{*}{\wdbc{}}}
& \lime{} & $.340$ & $453$ & $.460$ & $.516$ & $.650$ \\
& \calime{} & $\textbf{.298}$ & $\textbf{400}$ & $\textbf{.250}$ & $.244$ & $\textbf{.726}$ \\
& \textsc{f-lime} & $.338$ & $440$ & $.380$ & $\textbf{.248}$ & $.715$ \\
& \textsc{d-lime} & $.393$ & $490$ & $.521$ & $.251$ & $.705$ \\
\end{tabular}}
\end{table}
%

\section{Conclusion}
\label{sec:conclusion}
We have presented the first proposal in the research area of post-hoc local model-agnostic explanation methods that \textit{discovers} and \textit{incorporates} causal relationships in the explanation extraction process. 
In particular, we have used \gencda{} to sample synthetic data accounting for causal relationships.
Empirical results suggest that \calime{} can overcome the weaknesses of \lime{} concerning both the stability of the explanations and fidelity in mimicking the black-box.

From an application perspective, there is a growing demand for trustworthy and transparent AI approaches in high-impact domains such as financial services or healthcare. 
For instance, in medicine, this need arises from their possible applicability in many different areas, such as diagnostics and decision-making, drug discovery, therapy planning, patient monitoring, and risk management.
In these scenarios, due to mapping of explainability with causality, the exploitation of \calime{} will allow users to make an informed decision on whether or not to rely on the system decisions and, consequently, strengthen their trust in it. 

A limitation of our proposal is that adopting \gencda{} that in turn is based on \ncda{}, \calime{} can only work on datasets composed of continuous features. We need to rely on CD approaches to simultaneously account for heterogeneous continuous and categorical datasets to overcome this drawback.
For future research direction, it would be interesting to employ \gencda{} and, in general, the knowledge about causal relationships in the explanation extraction process of other model-agnostic explainers like \textsc{SHAP}~\cite{lundberg2017unified} or \textsc{lore}~\cite{guidotti2019factual}. Indeed, the \calime{} framework can be plugged into any model-agnostic explainer.
Finally, to completely cover \lime{} applicability, we would like to study to which extent it is possible to employ causality awareness on data types different from tabular data, such as images and time series.

{
\small
\bigskip
\textbf{Acknowledgments.}
This work is partially supported by the EU NextGenerationEU program under the funding schemes PNRR-PE-AI FAIR (Future Artificial Intelligence Research), PNRR-SoBigData.it - Strengthening the Italian RI for Social Mining and Big Data Analytics - Prot. IR0000013, H2020-INFRAIA-2019-1: Res. Infr. G.A. 871042 \emph{SoBigData++}, G.A. 761758 \emph{Humane AI}, G.A. 952215 \emph{TAILOR}, ERC-2018-ADG G.A. 834756 \textit{XAI}, and CHIST-ERA-19-XAI-010 SAI.
}

\bibliography{biblio}
\bibliographystyle{unsrt}
\end{document}